\pdfoutput=1
\documentclass[11pt]{article}

\usepackage[preprint]{acl}

\usepackage{times}
\usepackage{latexsym}
\usepackage[T1]{fontenc}
\usepackage[utf8]{inputenc}
\usepackage{microtype}
\usepackage{inconsolata}
\usepackage{amsmath}
\usepackage{amssymb}
\usepackage{amsthm}
\usepackage{graphicx}
\usepackage{mathtools}
\usepackage{listings}
\usepackage{tikz}
\usepackage{siunitx}
\usepackage{tipa}
\usepackage{float}
\usepackage{bbm}
\usepackage{breqn}
\usepackage{utils}
\usepackage{tikz-dependency}
\usepackage{booktabs}
\usepackage{hyperref}
\usepackage{linguex}
\usepackage{multirow}
\usepackage{seqsplit}
\usepackage{tabularx}
\usepackage{pifont}
\usepackage{array}
\usepackage[shortlabels]{enumitem}
\usepackage[normalem]{ulem}
\useunder{\uline}{\ul}{}

\usepackage{xcolor}

\usepackage{caption}
\usepackage{subcaption}
\usepackage{linguex}

\title{Bears, all bears, and some bears.\\Language Constraints on Language Models' Inductive Inferences}

\author{Sriram Padmanabhan \quad Siyuan Song \quad Kanishka Misra\\The University of Texas at Austin\\\texttt{\{srirampadmanabhan, siyuansong, kmisra\}@utexas.edu}}

\begin{document}
\maketitle
\begin{abstract}
Language places subtle constraints on how we make inductive inferences. Developmental evidence by \citet{gelman2002children} has shown children (4 years and older) to differentiate among generic statements (``\textit{Bears} are daxable''), universally quantified NPs (``\textit{all bears} are daxable'') and indefinite plural NPs (``\textit{some bears} are daxable'') in extending novel properties to a specific member (\textit{all} \textgreater{} \textit{generics} \textgreater{} \textit{some}), suggesting that they represent these types of propositions differently. We test if these subtle differences arise in general purpose statistical learners like Vision Language Models, by replicating the original experiment. On tasking them through a series of precondition tests (robust identification of categories in images and sensitivities to \textit{all} and \textit{some}), followed by the original experiment, we find behavioral alignment between models and humans. Post-hoc analyses on their representations revealed that these differences are organized based on inductive constraints and not surface-form differences.
\end{abstract}

\section{Introduction}
\label{sec:intro}

A hallmark of human cognition is our ability to make inductive inferences about categories \citep{osherson1990category, murphy2004big}. We can readily extend immediately available information along different categorical structures---e.g., on learning that \textit{robins} have a novel property, we might extend it to other \textit{birds}. An active area of research in cognitive science is to understand exactly how these inferences are constrained \citep{gelman2002children, hollander2002children, cimpian2008preschool}. For instance, a goldfish is a pet, a fish, an animal, and a living thing, and so deciding the extent to which properties generalize beyond available evidence, to other categories, is a non-trivial task---one which we master anyway. 

Language has been posited to play a particularly indispensable role in constraining category-based inductive inferences \citep{prasada2000acquiring, gelman2004learning}. One particular way in which language constrains inductive inferences is by modulating the scope of a proposition---e.g., \cref{ex:all} expresses a completely different inductive constraint (i.e., the extent to which it will generalize) about the novel property ``\textit{has the T9 hormone}'' than does \cref{ex:some}. 

\ex. \label{ex:example}
\a. \label{ex:all} \textit{All bears} have the T9 hormone. \textsc{[all]}
\b. \label{ex:some} \textit{Some bears} have the T9 hormone. \textsc{[some]}
\c. \label{ex:generic} \textit{Bears} have the T9 hormone. \textsc{[generic]}

\begin{figure}[!t]
    \centering
    \includegraphics[width=\columnwidth]{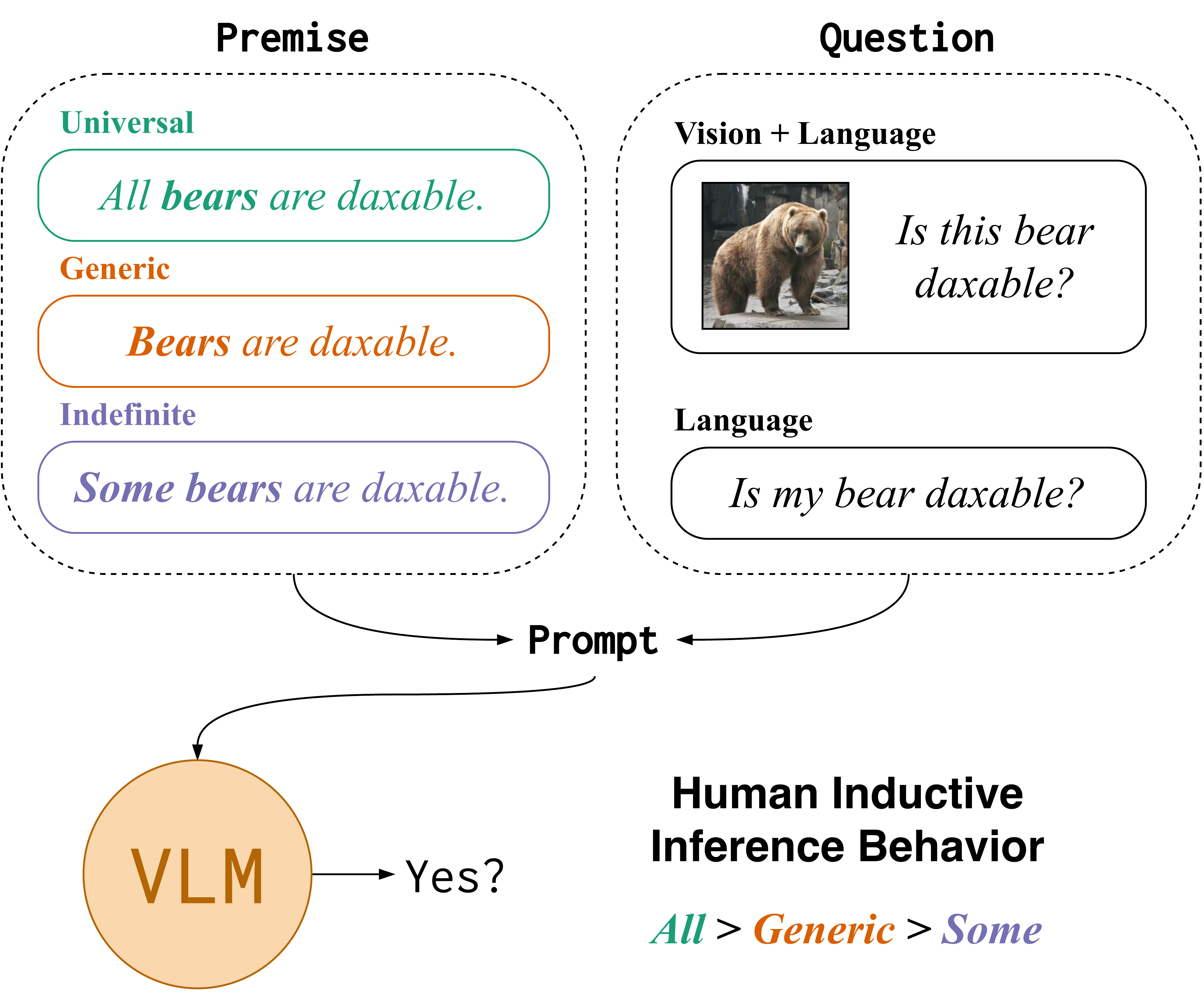}
    \caption{We study how the form of a proposition constrains a model's inductive inferences. We compare between universal quantifiers (\textit{all}), generics (\textit{bare plurals}), and indefinite quantifiers (\textit{some}). Humans represent them differently and show consistent graded effects in their inductive inferences \citep{gelman2002children}.
    }
    % An analogous experiment on humans \citep{gelman2002children} showed that they treat each separately, and show interesting graded pattern in their inductive inferences.}
    \label{fig:placeholder}
    \vspace{-1em}
\end{figure}

\noindent
While quantifiers like \textit{all} and \textit{some} are logically obvious in terms of the a priori expectations about how they might constrain a learner’s inductive inference, the status of statements like \cref{ex:generic} has been particularly puzzling from a theoretical standpoint. These statements---called \textit{generics} \citep{leslie2008generics}---denote law-like \textit{generalizations} about kinds, and are robust to counter-examples (encountering an unfriendly dog does little to alter one's generic knowledge that ``\textit{dogs are friendly}''). Generics are notoriously difficult to detect \citep{gelman2004learning}, especially because unlike quantifiers like \textit{all} and \textit{some}, generics are never explicitly marked in any known language. Simply treating bare plurals as generics is insufficient (e.g., they can also be denoted using indefinite articles---\textit{a goldfish has bad memory}) and sometimes incorrect (e.g., \textit{Mosquitoes are torturing me}). Nevertheless, children can produce generic sentences before more theoretically and logically tractable ones involving quantifiers like \textit{all} \citep{leslie2008generics}. Experimentally, \citet{gelman2002children} showed children (4 years old) and adults to represent generics like \cref{ex:generic} differently than they did \textit{all} and \textit{some} in their inductive behavior. Participants maximally extended a property to a newly encountered bear when told that \textit{all bears have a property}, followed by the bare plural \textit{bears have a property}, followed by \textit{some bears have a property}---i.e., their generalization behavior consistently  showed the following pattern: \textsc{all \textgreater{} generic \textgreater{} some}.

The puzzle of generics and its relation to quantifiers like \textit{all} and \textit{some} is quite relevant to modern artificial intelligence models like (vision) language models (VLMs). Apart from being central to the study of inductive reasoning in humans (a core property of our intelligence), the differences in the three kinds of propositions also allows us to shed light on how language \textit{form} ({all/some/generic}) interplays with language \textit{function} (inductive behavior) within these models. This is particularly relevant in light of recent positions about the ability of such models to grasp the nuance of meaning \citep{bender-koller-2020-climbing, mahowald2024dissociating}. 
Therefore in this paper, \textbf{we investigate the extent to which VLMs distinguish between \textsc{all}, \textsc{some}, and \textsc{generics} in their inductive inferences}. We do so by replicating \citet{gelman2002children}'s experiment on VLMs.\footnote{Our reason for using VLMs as opposed to LMs is because the original experiment by \citet{gelman2002children} was multimodal, and involved pictures of animals in its stimuli.} We provide models with premises consisting of statements like those in \Cref{ex:example} that attribute a property (\textit{have the T9 hormone}) to categories (\textit{bear}) and then test if they extend the property to a specific member of the category, presented to them as an image (like in the original experiment) or using other linguistic cues (e.g., \textit{my bear}). 

Before analyzing models' inductive reasoning behavior, we first test if they satisfy basic presuppositions to our research question, by tasking them with the same pretests as in \citet{gelman2002children}. This is especially important in order to make ``species-fair'' comparisons between models and humans \citep{firestone2020performance}. In particular, we test for two presuppositions. First, the models must recognize that their input image consists of the category that the premise and the subsequent question focuses on. For this, we test their ability to robustly identify categories in multiple object-centric images. Second, the models should capture the basic linguistic properties of \textit{all} and \textit{some}---both of which are theoretically and logically more tractable than generics. For this, we create a \textbf{new developmentally inspired benchmark} that tests if models can answer questions about a context presented to them in either modality (language vs. vision) targeting \textit{all} and \textit{some}---e.g., if the model says `yes' to the question ``Are \textit{all} blocks on the table \textit{blue}?'', pertaining to an image consisting of five blue blocks on a table and nothing else. Both these tests mimicked the pre-tests conducted by \citet{gelman2002children}, and the \textit{all} and \textit{some} test in particular was also inspired by a classic experiment testing for quantifier knowledge in children of ages 3--4 \citep{smith1980quantifiers}. We ran these tests on 10 different VLMs, finding at least two that robustly satisfied both these presuppositions.

Turning to our inductive behavior results, we found both VLMs from the previous experiments to show \textbf{qualitatively similar patterns of inductive reasoning as humans}. That is, they were more likely to generalize a novel property to a category when the premise contained a universal quantifier \textit{all}, followed by the bare plural \textit{generic}, followed by the indefinite quantifier \textit{some}. Our post-hoc analyses on models' vector representations of only the premises, without any task context, revealed that these proposition types occupied different regions in the models' low-dimensional representational space. This was found to hold even when we added stimuli that expressed the same proposition but differed in their surface forms to the original stimuli---e.g., adding \textit{every bear} to pair with \textit{all bears}. That is, models organized propositions \textbf{in terms of similarities in their inductive constraints, in a manner that cannot be explained by surface-form similarities/differences alone.}

Overall, our experiments and results contribute to the broader literature surrounding (V)LMs and their conceptual knowledge. Existing work on models' inductive reasoning capacities has largely focused on the effect of the conceptual content of the premises \citep{misra2021language, misra2022property, han2024inductive, bhatia2023inductive}. For instance, \citet{han2024inductive} use stimuli from \citet{osherson1990category}, which does not contain any variation in terms of scope-modifying cues.
We contribute to this body of work by narrowing our focus on the constraints imposed by various surface form changes, keeping the conceptual content the same (i.e., same concepts). 
Furthermore, our work also complements recent work on generics and (V)LMs. For instance, recent work has focused on the extent to which these models are able to reason \textit{about} generics in the context of their interplay with exceptions \citep{allaway-etal-2023-penguins, allaway2024exceptions, frank-allaway-2025-visage}. While generics have been explicitly compared against quantifiers \citep{ralethe-buys-2022-generic, collacciani-etal-2024-quantifying, cilleruelo-etal-2025-generics}, no work has investigated the differences between generics and quantifiers in terms of their inductive constraints, which is an important component of their meaning.
Finally, an unintended
contribution of our work is our developmentally inspired benchmark focusing on models' behavior on \textit{all} and \textit{some}, which we release in both the language and the vision+language modalities, and is the first of its kind, to the best of our knowledge---with previous work instead focusing on vague quantifiers \citep{wong2025vaquum}. This joins other developmentally inspired benchmarks for VLMs \citep{tan2024devbench, yiu2025kiva}. Our code is available at \url{https://github.com/kanishkamisra/inductive-constraints.}

\section{Models and measures}

\paragraph{Models} Our experiments use stimuli that span both the vision and text modalities, and therefore we conduct our analyses on Vision Language Models. We test on 10 different models: Idefics3-8B~\citep{laurençon2024building}, LLaVA-1.5 7B, LLaVA-v1.6-Mistral 7B~\citep{liu2023llava, liu2023improvedllava}, LLaVA-OneVision~\citep{li2025llavaonevision}, Molmo 7B~\citep{deitke2025molmo}, Qwen2.5-VL 7B Instruct~\citep{qwen2.5-VL} and Qwen3-VL Instruct 2/4/8B~\citep{qwen3technicalreport}, and SmolVLM Instruct~\citep[][2.2B]{marafioti2025smolvlm}. Models were accessed using the \texttt{minicons} library \citep{misra2022minicons}. Appendix \ref{sec:metadata} shows details of the VLMs.

\paragraph{Measures} Our stimuli in all three subsequent experiments involve a polar question, and expect models to answer with Yes or No. To account for surface form variation, we follow \citet{rodriguez-etal-2025-characterizing} and take into account space prefixing as well as both upper and lower cased versions of Yes and No into our measures. 
Given context $C$ and a question $Q$, with $\textsf{YES} = \{\texttt{`Yes'}, \texttt{`yes'}, \texttt{` Yes'}, \texttt{` yes'}\}$, and $\textsf{NO} = \{\texttt{`No'}, \texttt{`no'}, \texttt{` No'}, \texttt{` no'}\}$ we compute the relative probability of Yes as:
\begin{align}
    % \frac{\sum_{l\in \textsf{YES}} p_{\textsf{LM}}(l\mid Q, C)}{\sum_{l\in \textsf{YES}} p_{\textsf{LM}}(l\mid Q, C) + \sum_{l\in \textsf{NO}} p_{\textsf{LM}}(l\mid Q, C)},
    p_{\texttt{rel}}(\texttt{Yes}) = \frac{\sum_{l\in \textsf{YES}} p_{\textsf{LM}}(l\mid Q, C)}{\sum_{l\in \textsf{YES} \cup \textsf{NO}} p_{\textsf{LM}}(l\mid Q, C)},
\end{align}

\section{Experiment 1: Category Identification}
\label{sec:exp1}

Our first experiment targets VLMs' ability to robustly detect the presence (or absence) of categories in visual environments. This is an important precondition for our inductive generalization experiment, since a substantial part of that experiment involves testing models' generalization of a novel property to a category indicated using an image. We measure a model's robustness in category identification using two considerations: 1) by sampling multiple different `negative' categories (ones that do not appear in the image) for each instance, ensuring that the model not only detects the presence of a given category, but also the absence of those not in the image; and 2) by measuring accuracy as a function of multiple images, and only `rewarding' models iff. they are able to correctly perform category identification for \textit{all} images.

\begin{figure}[!t]
    \centering
    \includegraphics[width=\linewidth]{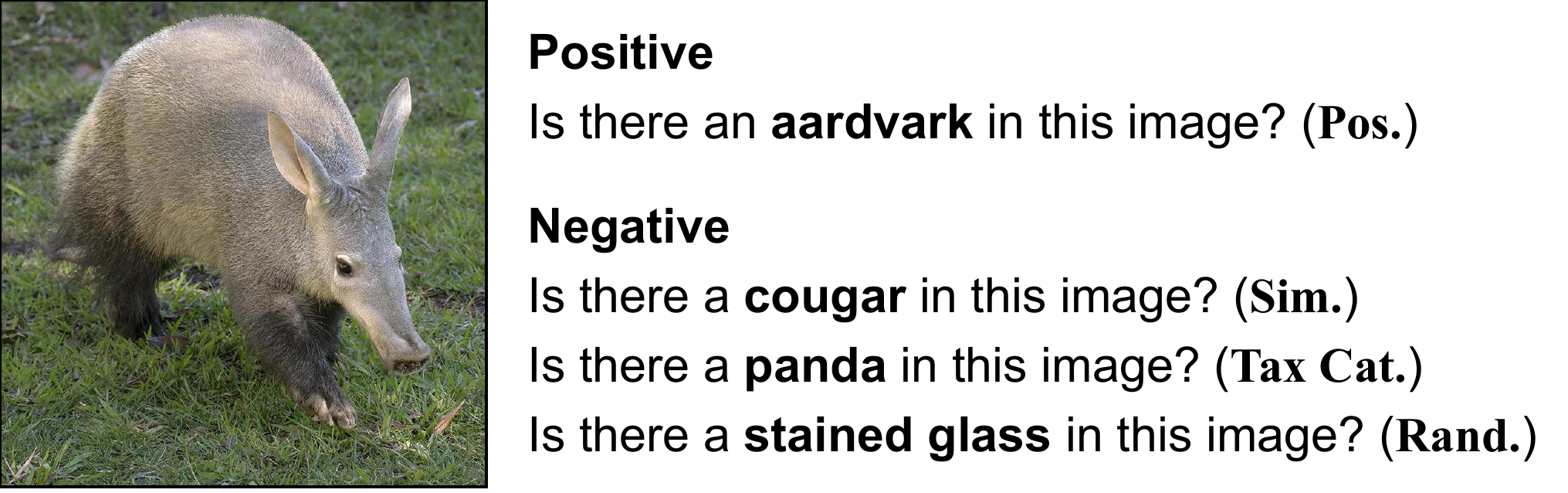}
    \caption{An instance from our category identification experiment. Each positive sample is associated with three separate types of negative samples.}
    \label{fig:category-identification-experiment}
    \vspace{-1em}
\end{figure}

\subsection{Data} 
Our images are sourced from THINGS~\citep{hebart2019things}, a resource of high quality, entity centric images collected for psychological experiments. THINGS consists of images across 1854 categories, with each category being paired with 10-20 images. 
We use a subset of 1222 categories from the database, and sample 5 images per category to measure image-level robustness. We use polar questions as our stimuli, with template ``\texttt{Is there a [CATEGORY] in this image? Answer with Yes or No.}''
Our positive samples (\textbf{Pos.}) are constructed by substituting \texttt{[CATEGORY]} with the appropriate lexical item as provided by THINGS. 
Following \citet{misra-etal-2023-comps} and \citet{qin2025visionandlanguage}, we pair each positive sample with multiple different negative samples, each coming from a different knowledge source (similarity, taxonomic relations, etc.). We consider three different sources:

\paragraph{SPoSE Similarity (Sim.)} Our first class of negative samples uses SPoSE embeddings \citep{zheng2019revealing, THINGSdata}, a vector space model built using the THINGS images, which provides a high-fidelity estimate for visual similarity judgments from humans \citep[Spearman's correlation of 0.94, see][]{kaniuth2024high}. For each target concept, we perform weighted random sampling of negative concepts, with the weights proportional to the SPoSE similarity between concepts.

\paragraph{Sampling from taxonomic category (Tax Cat.)} Our second class of negative samples comes from randomly sampling members of the smallest taxonomic category/hypernym that the target concept belongs to, as provided by \citet{THINGSdata} in the THINGS database. For instance, if \textit{sparrow} is our target, then we sample from the \textit{bird} category, expecting concepts like \textit{crow}, \textit{eagle}, etc. This strategy captures taxonomically related concepts. 

\paragraph{Random sampling (Rand.)} Finally our third class of negative samples comes from a simple random sampling over the other THINGS categories.

Since there are 5 unique images per positive concept, and 3 total negative samples, we end up with a total of 24,440 image, question pairs.

\begin{table}[!t]
\centering
\resizebox{\linewidth}{!}{%
\begin{tabular}{lcccc|c}
\toprule
\textbf{Model} & \textbf{Pos.} & \textbf{Sim.} & \textbf{\begin{tabular}[c]{@{}c@{}}Tax\\ Cat.\end{tabular}} & \textbf{Rand.} & \textbf{Joint} \\ \midrule
Llava-1.5-7B & 97.5 & 26.6 & 52.2 & 78.0 & 15.0\\
Llava-OV-7B & 35.8 & 58.9 & 79.0 & 90.0 & 17.4\\
Llava-1.6-7B & 91.0 & 40.5 & 69.5 & 90.5 & 26.5\\
SmolVLM & 80.1 & 58.0 & 82.4 & 95.0 & 37.8\\
Idefics3-8B & \textbf{98.4} & 51.8 & 78.4 & 93.8 & 40.1\\
Molmo-7B & 85.0 & 71.6 & 88.0 & 94.8 & 52.9\\
Qwen3-VL-2B & 95.2 & 71.8 & 89.1 & 96.1 & 60.0\\
Qwen3-VL-4B & {\ul 98.3} & 72.4 & 90.5 & {\ul 97.3} & 63.8\\
Qwen3-VL-8B & 95.4 & {\ul 75.4} & {\ul 91.4} & 97.0 & {\ul 65.2}\\
Qwen2.5-VL-7B & 93.3 & \textbf{77.5} & \textbf{92.1} & \textbf{97.5} & \textbf{66.2}\\ \bottomrule
\end{tabular}%
}
\caption{Category Identification Results. Accuracies of VLMs on positive samples (\textbf{Pos.}), and the three negative sample subsets (\textbf{Sim.} = Similarity-based; \textbf{Tax. Cat.} = Sampling from same taxonomic category; \textbf{Rand.} = Random sampling). \textbf{Joint} indicates the proportion of time a model correctly answered all 20 questions for a given category. \textbf{Bold} and \underline{underline} indicate best and second-best performance, respectively.}
\label{tab:category-id-results}
\vspace{-1em}
\end{table}

\begin{table*}[!t]
\centering
\resizebox{0.75\textwidth}{!}{%
\begin{tabular}{@{}lllc@{}}
\toprule
\textbf{Condition} & \textbf{Vision + Language} & \textbf{Language} & \textbf{Answer} \\ \midrule
All/All/All & \begin{minipage}{.3\textwidth}
      \includegraphics[width=\linewidth]{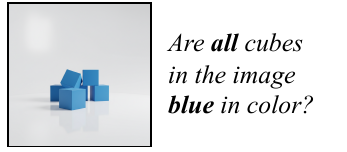}
    \end{minipage} & \textit{\begin{tabular}[c]{@{}l@{}}There are 20 blocks on a table.\\ 20 of them are \textbf{red}. \\ Are \textbf{all} of the blocks on the table \textbf{red}?\end{tabular}} & Yes \\ \midrule
All/All/None & \begin{minipage}{.3\textwidth}
      \includegraphics[width=\linewidth]{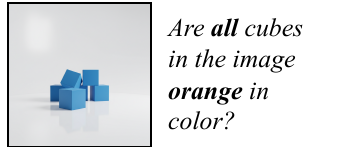}
    \end{minipage} & \textit{\begin{tabular}[c]{@{}l@{}}There are 20 blocks on a table.\\ 20 of them are \textbf{red}.\\ Are \textbf{all} of the blocks on the table \textbf{blue}?\end{tabular}} & No \\ \midrule
Some/All/Maj. & \begin{minipage}{.3\textwidth}
      \includegraphics[width=\linewidth]{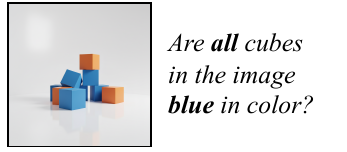}
    \end{minipage} & \textit{\begin{tabular}[c]{@{}l@{}}There are 20 blocks on a table.\\ 12 of them are \textbf{red} and 8 of them are \textbf{blue}.\\ Are \textbf{all} of the blocks on the table \textbf{red}?\end{tabular}} & No \\ \midrule
Some/All/Min. & \begin{minipage}{.3\textwidth}
      \includegraphics[width=\linewidth]{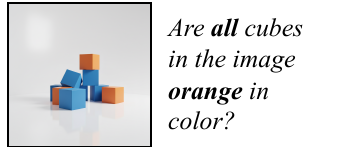}
    \end{minipage} & \textit{\begin{tabular}[c]{@{}l@{}}There are 20 blocks on a table.\\ 12 of them are \textbf{red} and 8 of them are \textbf{blue}.\\ Are \textbf{all} of the blocks on the table \textbf{blue}?\end{tabular}} & No \\ \midrule
Some/Some/Maj. & \begin{minipage}{.3\textwidth}
      \includegraphics[width=\linewidth]{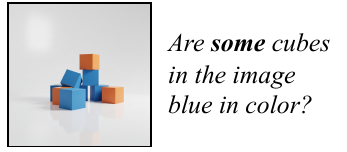}
    \end{minipage} & \textit{\begin{tabular}[c]{@{}l@{}}There are 20 blocks on a table.\\ 12 of them are \textbf{red} and 8 of them are \textbf{blue}.\\ Are \textbf{some} of the blocks on the table \textbf{red}?\end{tabular}} & Yes \\ \midrule
Some/Some/Min. & \begin{minipage}{.3\textwidth}
      \includegraphics[width=\linewidth]{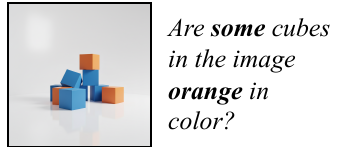}
    \end{minipage} & \textit{\begin{tabular}[c]{@{}l@{}}There are 20 blocks on a table.\\ 12 of them are \textbf{red} and 8 of them are \textbf{blue}.\\ Are \textbf{some} of the blocks on the table \textbf{blue}?\end{tabular}} & Yes \\ \bottomrule
\end{tabular}%
}
\caption{Stimuli for testing knowledge of \textit{all} and \textit{some}, across input modalities: 1) \textbf{Vision + Language} (\textit{N}=840), where questions target properties of entities in an image; and 2) \textbf{Language} (\textit{N}=900), where questions target properties of entities in a math word-problem. There are a total of 6 conditions, varying in the properties of the input context (All/Some), the question (All/Some), and the amount of entities to which the property in the question targets (All/None/Majority/Minority). \textbf{Note:} stimuli in the two modalities are \textit{not} matched and generated separately.}
\label{tab:all-some-stimuli}
\end{table*}

\subsection{Results and Analysis}

To recap our stimuli, each category is associated with 5 total images, and for each image, we have 4 types of questions---one positive question which asks for the presence of the category (expecting `Yes'), and three negatives, which ask for the presence of other category (expecting `No'). A competent model must be able to identify not only that the object in \textit{all} images (per category) is the target category but also that it is not some other category. This results in two types of measures. First, for each question type, we measure the average percentage of cases the model correctly answers the question correctly for \textit{all} images for a given category. We report this accuracy measure per question-type. Additionally, we also measure the average proportion of time the model correctly answers \textit{all} four questions for \textit{all} five images in the given category---i.e., all 20 questions associated with a given category. We refer to this latter measure as `\textbf{Joint}' accuracy. \Cref{tab:category-id-results} shows the accuracy measure across all question types as well as the Joint accuracy.

We see that VLMs achieved generally high accuracies on positive questions and randomly sampled negative questions. Their performance declined on negative questions constructed using concepts drawn from the same taxonomic category, though not by all that much. Finally, across all models, the lowest accuracy was observed on similarity-based negative samples. This insensitivity to similarity-based samples corroborates previous work on LMs' ability to predict features \citep{misra-etal-2023-comps}. Out of all evaluated VLMs, Qwen2.5-VL-7B attained the highest \textbf{Joint} accuracy across all four sample types, with Qwen3-VL-8B being a close second.

\section{Experiment 2: Behavioral Sensitivity to \textit{All} and \textit{Some}}

Our next experiment focuses on models' representation of the universal quantifier \textit{all} and the indefinite quantifier \textit{some}. We use this as a pretest to establish the minimal linguistic knowledge of the models to complete the task in our inductive inference experiments. 
This test specifically targets if models realize \textit{all} as a quantifier to denote a scenario where every item had an attribute (e.g., color) and \textit{some} as a quantifier where at least one item had the attribute. The test follows directly from \citet{gelman2002children}, as well as classic experimental work on children's ability to learn knowledge of quantifiers \citep{smith1980quantifiers}, who showed that children begin to learn the difference between \textit{all} and \textit{some} at age 3.
However, since there currently is no such dataset that allows us to test this directly, we resort to creating our own tightly controlled benchmark consisting of stimuli in both vision and textual modalities.

\subsection{Data}

We design separate stimuli across visual and textual modalities. Nonetheless, our stimuli in both modalities follow a general template consisting of a context (visual or textual) that presents objects with one or more properties, followed by a question that inquires about the properties of these objects. Our stimuli in a given modality vary in terms of the presented context (2 all, 4 some), target quantifier of the question (3 all 3 some), and the number of objects to which the property in the question applies to in the context (all/none/majority/minority). For instance, if the context is an image with  5 blue blocks and 3 orange blocks, and the question is \textit{Are all blocks in the image blue in color?}, then this would be coded as `Some/All/Majority', since the context shows a scenario where some objects have the property, the question targets all objects, and the property applies to a majority of the objects in the question. \Cref{tab:all-some-stimuli} shows detailed stimuli.

\paragraph{Vision + Language Stimuli} 
Due to a lack of existing datasets that test knowledge of quantifiers in VLMs in a controlled manner, we design our own Vision and Language dataset, inspired by \citet{gelman2002children} and \citet{smith1980quantifiers}. To this end, we prompt Gemini 2.5 Flash Image~\citep{comanici2025gemini} to first generate an image with a specified property that holds true for either all \textit{or} some objects (e.g., flowers in a vase), we then prompt it to edit the image by adding new objects such that the property now applied to the opposite quantifier (e.g., adding a few flowers outside the vase). This gives us our input context variation between \textit{all} and \textit{some}. To ensure the generated images are consistent with our instructions, we manually analyzed \textit{every} generation. Specifically, we prompted the model to generate 200 pairs of images, and then checked if 1) the pairs were minimally different (no visible difference in backgrounds), still maintained the all vs.~some distinction, and 3) did not add additional properties beyond what was specified. This resulted in 140 pairs in total. 
Appendix \ref{sec:all-some-generation} shows full details about our manual annotation.
We then paired these images with their respective questions, as described in the previous paragraph (6 different conditions), giving us a total of 840 stimuli. 
These stimuli mimic those used by \citet{gelman2002children}.

\paragraph{Language Stimuli} Our language stimuli include synthetically generated contexts posed as word-problems about objects and their properties. We generate 150 unique contexts differing in objects (\textit{N}=5, e.g., \textit{coins in a jar}, \textit{bottles in a crate}, etc.), property pairs (\textit{N}=6, e.g., \textit{red/blue in color}, \textit{contains milk/soda}, etc.), and amounts (\textit{N}=5, e.g., 500 total objects, with 300 having a given property and 200 having the other property). This results in a total of 900 stimuli (when combined with 6 different conditions). The example in Table 2 uses as its object `blocks on a table', as its property pair, `red/blue in color' and as its amount, 20 total objects---12 having the majority property (red) and 8 having the minority property (blue).

\begin{figure}[!t]
    \centering
    \includegraphics[width=\linewidth]{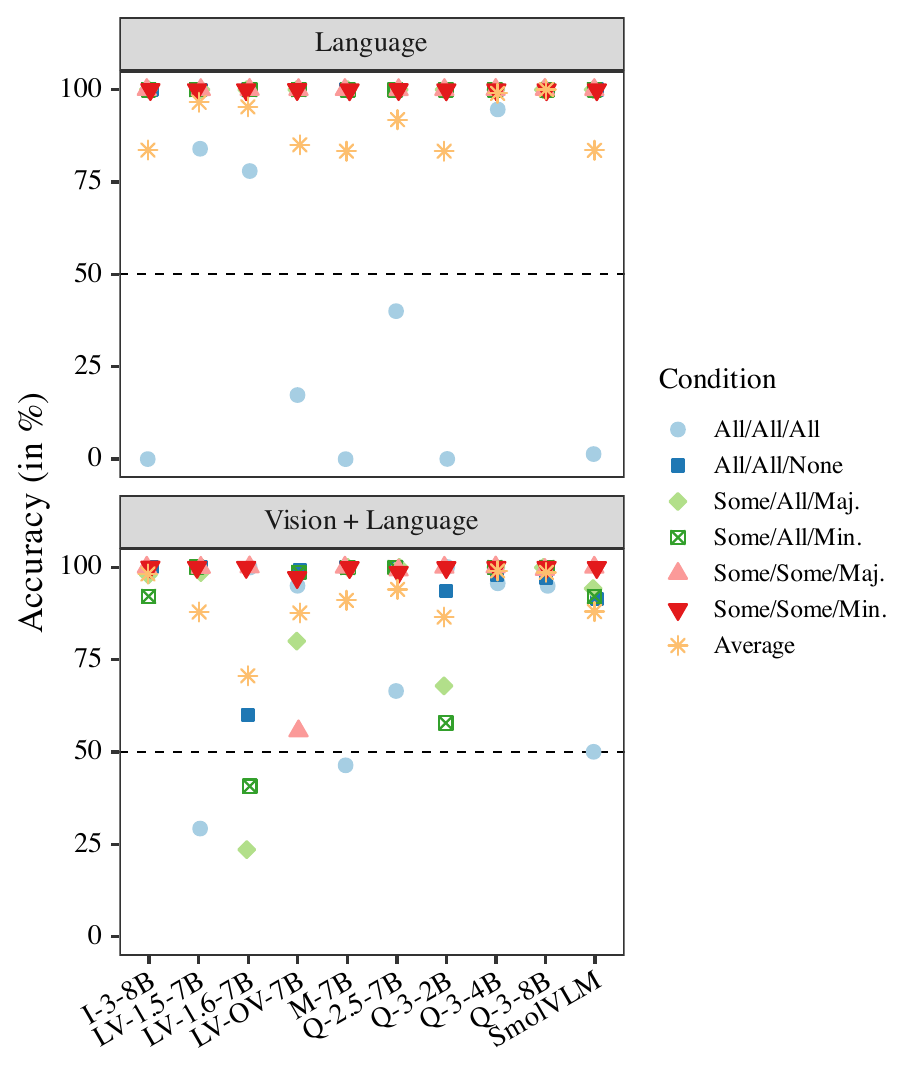}
    \caption{Accuracy of VLMs on \textit{all} and \textit{some}, across modalities.~Each point represents the accuracy of a model on a subset of the dataset, except for `Average', which denotes the average accuracy across subsets.}
    \label{fig:all-some}
    \vspace{-1em}
\end{figure}

\subsection{Results and Analysis}
 
\Cref{fig:all-some} shows the accuracy of all models across question-types for each condition, for both modalities.
Across both modalities, the VLMs generally achieved high accuracy on questions coded as Some/Some/*. In contrast, performance on Some/All/* questions exhibited notable modality-dependent variation. Specifically, all models answered these questions with near-perfect accuracy in the language-only setting, whereas several models were noticeably worse when presented with vision–language inputs, highlighting difficulties that VLMs often have reasoning about ``All'' and ``Some'' in the visual domain. More pronounced differences emerged for questions coded as All/All/All---a large subset of models performed poorly on these questions, often at or below chance level (especially in the language-only domain), despite maintaining high accuracy on All/All/None questions. This pattern suggests a systematic bias toward negative (“no”) responses when models are asked to verify whether a property holds universally across all objects in a scene. Among the evaluated models, only Qwen3 4B and 8B demonstrated near-perfect accuracy across all conditions and modalities.

\section{Experiment 3: Constraints on Inductive Inference}

Having shown that there exist VLMs (Qwen3-VL 4B and 8B) that can robustly detect categories in images, \textit{as well as} show desirable behavior in distinguishing between \textit{all} and \textit{some}, across modalities, we now turn to our inductive generalization experiments. Here, we test if these VLMs distinguish between propositions that involve explicit quantifiers---\textit{all} and \textit{some}---from those involving \textit{generics}. Following \citet{gelman2002children}, we provide models with premises that attribute a property to a category, and varying in the type of proposition (\textit{all/some/generic}) and then ask if a specific member of the category has that property. 
We then test if VLMs, like humans, show distinct behavior on \textit{all} vs. \textit{generic} vs. \textit{some} \citep{gelman2002children, hollander2002children}. More specifically, if they show the same qualitative pattern of generalizing the property in the premise to the category member most for \textit{all}, followed by \textit{generic}, followed by \textit{some}. Since the 4B and 8B versions of Qwen3-VL were the only models that \textit{robustly} satisfied our preconditions, we use them for this experiment.

\subsection{Data}

\paragraph{Category selection}  
We only consider categories for which both models of interest were able to correctly answer all 20 questions (see \Cref{sec:exp1}). Further, most research on generics has largely focused on generic statements about animate concepts---this is presumably because a majority of generics are stated \textit{for} animates \citep[and specifically, animals---see][]{gelman2008generic}. To test if the animacy of the premise category has an effect, we sample 50 animate and inanimate categories each ($N_{\textit{total}}$=100). 
 
\paragraph{Stimuli} 
Our stimuli have two components: a premise and a question. The premise, which follows the same format across our experiments on different modalities, expresses a proposition attributing a property to a category, with the constraints of the scope being modulated by \textit{all}, \textit{some}, or \textit{generic}. To ensure that the influence of the models' parametric knowledge on their inductive generalization behavior is minimal, we create \textit{novel} properties, using both nonce words (e.g., is/are daxable, has/have feps, etc.) and more `real' sounding properties (e.g., has the T9 hormone, is made of bergentium). In total, we have 5 nonce word-based properties and 5 real-sounding properties. Further, we include 4 different prompt variations of expressing the premise (e.g., \textit{``Given that \texttt{\{premise\}}''}, etc.). This results in a total of 40 different surface form variations---see Appendix \ref{appsec:surface-form} for full details about the surface form variation in our stimuli. The `question' component of the stimuli asks if the property applies to a specific member of the category. For the vision and language stimuli, we do this by using an image followed by a demonstrative---e.g., \textit{\texttt{[image]} Given that all bears have the T9 hormone, does this bear have the T9 hormone?}---following \citet{gelman2002children}. For our language stimuli, we use the possessive determiner ``\textit{my}''---e.g., \textit{Given that all bears have the T9 hormone, does my bear have the T9 hormone?} All in all, we have 40 surface form variations, 100 categories, and 3 types of premises (all vs. some vs. generic). This gives us 12,000 language stimuli, and 60,000 vision and language stimuli (accounting for 5 images per category).

\subsection{Results and Analysis}
\label{sec:main-exp-results}

\paragraph{Main results} To evaluate whether VLMs distinguish between \textit{all}, \textit{some}, and \textit{generic} in their inductive generalization, we compute the relative probability of a ``yes'' response, $p_{\texttt{rel}}(\text{Yes})$, as defined in Eq. (1), for each stimulus. We then average $p_{\texttt{rel}}(\text{Yes})$ across stimuli corresponding to each quantifier category. Figure 4 reports these averages, stratified by modality and animacy, for each model. In general, models exhibit the highest propensity to attribute a property to an individual category member when the property holds for \textit{all} members of the category, followed by \textit{generic}, followed by \textit{some}. 
This was also reflected in a linear mixed effects model analysis, predicting $p_{\texttt{rel}}(\text{Yes})$ by using premise type, animacy, and modality as fixed effects, and concept and prompt template as random effects ($p < .001$ for both VLMs).
This trend is observed across both modalities, implying that the presence/absence of an image of a member of a category does not affect the models' qualitative sensitivity to all/some/generic. This indicates a more general representation of \textit{all}, \textit{some}, and \textit{generic}. Overall, the $p_{\texttt{rel}}(\text{Yes})$ in the generics premise is consistently lower for inanimate categories than for animate categories, which aligns with the fact that generics are typically used in the context of animate objects \citep{gelman2008generic}. To our knowledge, humans were never tested in a manner similar to that of \citep{gelman2002children} for inanimate categories, leaving this as a possible direction for future work. Overall, VLMs seem to show qualitatively similar results has humans \citep{gelman2002children, hollander2002children}, with both systems showing the pattern \textsc{all > generics > some}.

\begin{figure}[!t]
    \centering
    \includegraphics[width=\linewidth]{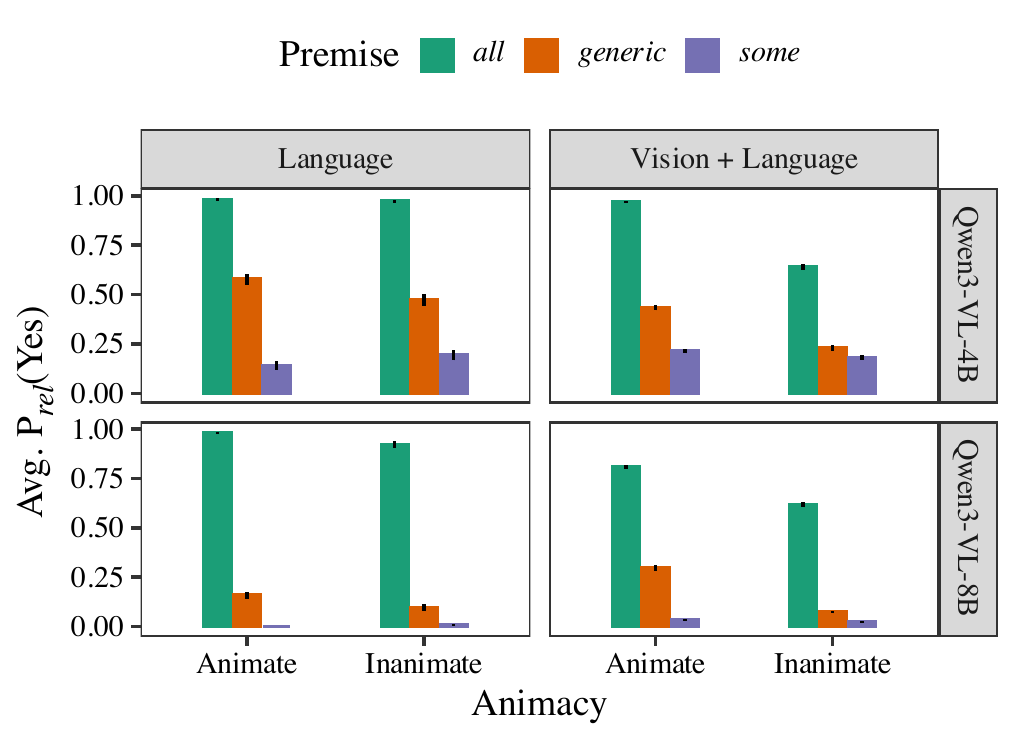}
    \caption{Avg. probabilities of extending the property from the premise to the specific instance in the question (i.e., predicting \texttt{Yes}) in Qwen3-VL models (4B and 8B) for animate and inanimate categories, across both modalities, and premise types (\textit{all} vs. \textit{generics} vs. \textit{some}). Error bars denote 95\% confidence intervals, over prompt variations (templates and properties).}
    \label{fig:placeholder}
    \vspace{-1em}
\end{figure}

\begin{figure*}[!t]
    \centering
    \includegraphics[width=\linewidth]{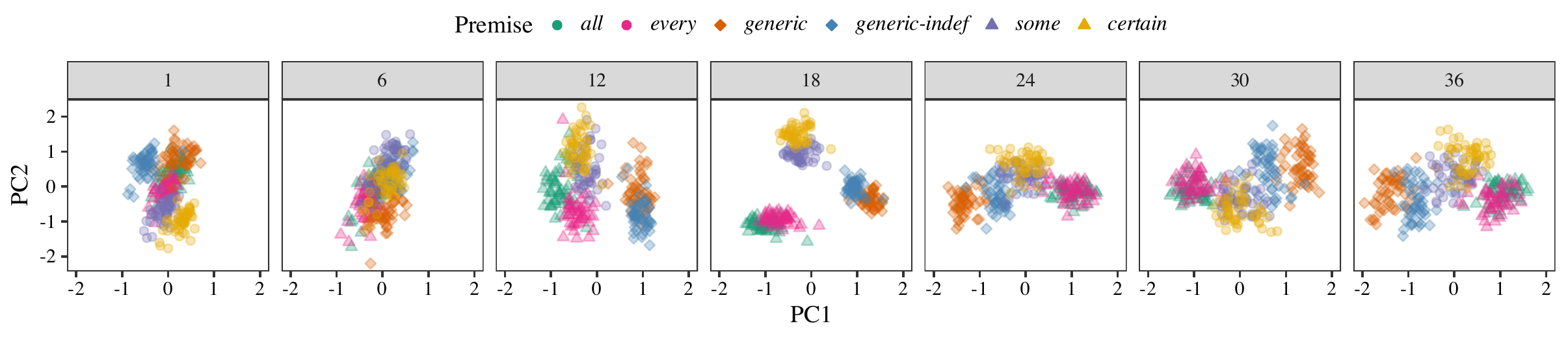}
    \caption{First two Principal Components of the last hidden state representations in selected layers of Qwen3-VL-8B for stimuli that attribute properties to categories and vary in their scope, as modulated by quantifiers (all/every vs. certain/some) or generics (bare plural/indefinite). Results for all layers are shown in \Cref{fig:pca-full}.}
    \label{fig:qwen3-8b-pca}
    \vspace{-1em}
\end{figure*}

\paragraph{Post-hoc representation analysis} Our behavioral results show that Qwen3 VLMs show distinct patterns of inductive generalization when given premises that express universal, generic, or indefinite propositions; especially for animate kinds. 
To what extent do these distinctions arise in the internal representations of the models? Do the Qwen3 VLMs organize propositions in a manner that is consistent with their inductive behavior? To answer these questions, we investigate low-dimensional representations of the models' hidden states on sentence stimuli that \textit{only} contain the specific proposition, without any task context. By removing task information, we ensure that the patterning of the models' representation is not biased to mimic their next word-probabilities, which we have already established to be distinguished. 
A potential confounding factor in this analysis could be that the three premise types are distinguished by surface forms (all vs. some vs. bare plurals), and therefore, differences in the model's internal representations could simply be explained by surface form differences. In order to combat this, we add a completely different kind of proposition for each premise type. 
Specifically, we add \textit{every} followed by a singular noun denoting the category to pair with \textit{all}, \textit{certain} to pair with \textit{some}, and an indefinite article `a/an' and a singular noun denoting the category to pair with the bare plural \textit{generic}, as shown in \cref{ex:additional}:

\ex. \label{ex:additional}
\a. \textit{Every} bear is daxable. \textsc{[all]}
\b. \textit{Certain} bears are daxable. \textsc{[some]}
\c. \textit{A bear} is daxable. \textsc{[generic]}

Finally, we restrict our examples to animate categories, and nonce-properties ($N$=250 stimuli per proposition type, and 6 propositions), and extract the models' hidden state at the final token position across all layers. We then perform principal components analysis (PCA) on these representations. \Cref{fig:qwen3-8b-pca} visualizes the first two principal components for layers 1, 6, 12, 18, 24, 30, and 36 (the final layer) of Qwen3-VL 8B. 

If the representations are \textit{only} sensitive to surface form as opposed to the similarity in the propositions in terms of their inductive constraints, then we would see six different clusters of points, all far from each other. Instead, from \Cref{fig:qwen3-8b-pca}, we see that the representations are largely entangled (in low-dimensional spaces) in the earlier layers, and then gradually organize themselves in terms of their inductive constraints in the middle layers. For instance, in layer 18, we see maximal entanglement in the representations of near-synonymous propositions (\textit{every} with \textit{all}, \textit{certain} with \textit{some}, and indefinite \textit{generic} [\textit{generic-indef}] with bare plural \textit{generic}), as well as maximal separation between propositions with different meanings. This in fact starts to emerge at layer 13 (see \Cref{fig:pca-full} in the appendix). In later layers, while the separation among propositions with different meanings is further reduced, propositions similar in meaning are still closer to each other than they are to other propositions. We provide quantitative analysis of this effect in Appendix \ref{appsec:pca-full}. Overall, these results suggest that the propositions differing in their inductive constraints---as modulated by quantifiers or generics---are represented separately in VLMs' representational space, in a manner that cannot be explained by surface-form differences alone.

\section{Conclusion}

The \textit{form} of language can have systematic impact on its \textit{function}. We have presented an analysis of how this could be realized within VLMs, by focusing on how subtle linguistic cues that modulate the scope of a proposition affect these models' inductive inferences involving categories. Our findings suggest that VLMs represent propositions varying in these cues in ways that are similar to those in humans \citep{gelman2002children}. 
That is, there was general behavioral alignment between how \textit{all}, \textit{some}, and \textit{generics} constrain the inductive behavior between both systems (humans and models).
Overall, since generics in particular have been a great mystery to philosophers, cognitive scientists, and linguists \citep{gelman2004learning, leslie2008generics}, we are hopeful that further analyses on (V)LMs can help shed light on the underlying mechanisms for the acquisition and processing of generic statements \citep{allaway2024exceptions}.

\section*{Limitations}
This study is a foray into understanding the connection between generics and inductive inferences from a computational perspective. While we were able to establish behavioral alignment between VLMs and humans, our study has several limitations, which we enumerate below.

\paragraph{Lack of Causal Implications}
First, while we have shown generics to be represented separately from quantifiers like \textit{all} and \textit{some} in a manner that is not just determined by surface form differences, we have not established any causal implications of such representations. In general there have been cautions against exclusively concluding from from simpler low-dimensional analyses, since they may be biased towards certain kinds of features \citep{lampinen2025representation}. A more thorough investigation using causal interpretability methods \citep{geiger2024finding} could potentially alleviate this issue, though it is not entirely clear how one might causally localize regions of interest pertaining to the three types of propositions in a manner that does not encode differences between them---ideally we would want to discover this in an unsupervised manner.

\paragraph{Model Choice and `Cognitive Plausibility'}
Perhaps one of the biggest limitations of this work is that the VLMs we investigate have been trained on orders of magnitude more data than humans, which raises questions about the transfer of insights between models and humans. While we have refrained from making any claims about human congition from our analyses, especially due to this fact, we have enforced strict preconditions in order to select the right model to investigate and have remained faithful to the pretests of \citet{gelman2002children}, which this paper aims to replicate. At the same time, a recent working hypothesis, called the \textit{contravariance principle}, has emerged in the sub-community of researchers interested in establishing a transfer of insight between neural networks and human minds \citep{cao2024explanatory, cao2024contra, futrell2025linguistics}. According to this principle, if two systems seemingly solve the same problem that is \textit{hard}---the sense that solving it requires satisfying innumerable constraints---then their solutions are similar in ``functionally relevant'' ways, even though the systems themselves might differ from one another in ``functionally irrelevant ways'' \citep{cao2024contra}. Since generics in particular has been one of the hardest puzzles in the cognitive psychology of language and thought \citep{leslie2008generics}, there is some benefit to first establish that the two systems that we have concerned ourselves with (VLMs and Humans) converge at the same behavioral sensitivities. In this way, we have established which models are better animal models that we can then investigate more mechanistically in order to articulate novel predictions and hypotheses \citep{mccloskey1991networks, lakretz2021mechanisms, misra2024generating}.

\paragraph{Sample Sizes for All and Some}
Our benchmark for establishing behavioral sensitivity to \textit{all} and \textit{some}---inspired from developmental work though it may be---only has a few hundred instances, limited to a handful of attributes and objects. While the benchmark is not the main contribution of the paper and serves more as a pre-test for models, it would be worthwhile to increasing the benchmark's sample size and diversity, as well as include an entire host of quantifiers, in future work.

\paragraph{Model generated stimuli for All and Some}
Our \textit{all} and \textit{some} pretest involves both language and vision+language stimuli. While we generated our language stimuli ourselves, we had to rely on Gemini 2.5 \citep{comanici2025gemini} to generate image stimuli for our vision+language subset. The main reason behind this was that there were no controlled datasets that allowed us to test models' behavior on \textit{all} and \textit{some} in the same manner as \citet{smith1980quantifiers} and \citet{gelman2002children}. Furthermore, it is expensive---intractable, even---to find images in the wold that could serve as controlled stimuli that differ only minimally in the applicability of a quantifier. Finally, we held out stimuli towards the strictest of standards, and manually inspected each and every single generated image. We ended up rejecting about 30\% of the total generated image-pairs, and have disclosed full details in \Cref{sec:all-some-generation}. Our usage of Gemini to generate stimuli for this benchmark is another reason why our sample sizes are small (see above limitation).

\paragraph{Monolingual Investigation} Finally, our stimuli only focus on the English language, whereas generics have been a universal puzzle across languages, because they are never explicitly marked \citep{leslie2008generics}. Understanding them from a multi-lingual perspective would make for important future work.

\section*{Acknowledgments}

We are grateful to Susan Gelman and her group for their work on generics, kinds, and inductive inferences, which is the important foundational work we build on. We are also grateful to the THINGS initiative \citep{hebart2019things, THINGSdata} for making their data open and accessible.
K.M. is supported by the Donald D. Harrington Faculty Fellowship at UT Austin. The authors acknowledge the Texas Advanced Computing Center (TACC) at The University of Texas at Austin for providing computational resources that have contributed to the research results reported within this paper. URL: \url{http://www.tacc.utexas.edu}.

\bibliography{bibliography}

@article{misra2022minicons,
  title={minicons: Enabling flexible behavioral and representational analyses of transformer language models},
  author={Misra, Kanishka},
  journal={arXiv:2203.13112},
  year={2022}
}

@article{gelman2002children,
  title={Children's use of generics in inductive inferences},
  author={Gelman, Susan A and Star, Jon R and Flukes, Jonathan},
  journal={Journal of Cognition and Development},
  volume={3},
  number={2},
  pages={179--199},
  year={2002},
  publisher={Taylor \& Francis}
}

@article{hollander2002children,
  title={Children's interpretation of generic noun phrases.},
  author={Hollander, Michelle A and Gelman, Susan A and Star, Jon},
  journal={Developmental psychology},
  volume={38},
  number={6},
  pages={883},
  year={2002},
  publisher={American Psychological Association}
}

@inproceedings{zheng2019revealing,
  author       = {Charles Y. Zheng and
                  Francisco Pereira and
                  Chris I. Baker and
                  Martin N. Hebart},
  title        = {Revealing interpretable object representations from human behavior},
  booktitle    = {7th International Conference on Learning Representations, {ICLR} 2019,
                  New Orleans, LA, USA, May 6-9, 2019},
  publisher    = {OpenReview.net},
  year         = {2019},
  url          = {https://openreview.net/forum?id=ryxSrhC9KX},
  bibsource    = {dblp computer science bibliography, https://dblp.org}
}

@article{hebart2019things,
  title={THINGS: A database of 1,854 object concepts and more than 26,000 naturalistic object images},
  author={Hebart, Martin N and Dickter, Adam H and Kidder, Alexis and Kwok, Wan Y and Corriveau, Anna and Van Wicklin, Caitlin and Baker, Chris I},
  journal={PloS one},
  volume={14},
  number={10},
  pages={e0223792},
  year={2019},
  publisher={Public Library of Science San Francisco, CA USA}
}

@article {THINGSdata,
	article_type = {journal},
	title = {THINGS-data, a multimodal collection of large-scale datasets for investigating object representations in human brain and behavior},
	author = {Hebart, Martin N and Contier, Oliver and Teichmann, Lina and Rockter, Adam H and Zheng, Charles Y and Kidder, Alexis and Corriveau, Anna and Vaziri-Pashkam, Maryam and Baker, Chris I},
	editor = {Barense, Morgan},
	volume = 12,
	year = 2023,
	month = {feb},
	pub_date = {2023-02-27},
	pages = {e82580},
	citation = {eLife 2023;12:e82580},
	doi = {10.7554/eLife.82580},
	url = {https://doi.org/10.7554/eLife.82580},
	journal = {eLife},
	issn = {2050-084X},
	publisher = {eLife Sciences Publications, Ltd},
}

@article{kaniuth2024high,
  title={A high-throughput approach for the efficient prediction of perceived similarity of natural objects},
  author={Kaniuth, Philipp and Mahner, Florian P and Perkuhn, Jonas and Hebart, Martin N},
  journal={bioRxiv},
  pages={2024--06},
  year={2024},
  publisher={Cold Spring Harbor Laboratory}
}

@article{prasada2000acquiring,
  title={Acquiring generic knowledge},
  author={Prasada, Sandeep},
  journal={Trends in cognitive sciences},
  volume={4},
  number={2},
  pages={66--72},
  year={2000},
  publisher={Elsevier}
}

@article{li2025llavaonevision,
title={{LL}a{VA}-OneVision: Easy Visual Task Transfer},
author={Bo Li and Yuanhan Zhang and Dong Guo and Renrui Zhang and Feng Li and Hao Zhang and Kaichen Zhang and Peiyuan Zhang and Yanwei Li and Ziwei Liu and Chunyuan Li},
journal={Transactions on Machine Learning Research},
issn={2835-8856},
year={2025},
url={https://openreview.net/forum?id=zKv8qULV6n},
note={}
}

@article{smith1980quantifiers,
  title={Quantifiers and question answering in young children},
  author={Smith, Carol L},
  journal={Journal of experimental child psychology},
  volume={30},
  number={2},
  pages={191--205},
  year={1980},
  publisher={Elsevier}
}

@article{gelman2004learning,
  title={Learning words for kinds: Generic noun phrases in acquisition},
  author={Gelman, Susan A},
  journal={Weaving a lexicon},
  pages={445--484},
  year={2004}
}

@article{gelman2008generic,
  title={Generic language in parent-child conversations},
  author={Gelman, Susan A and Goetz, Peggy J and Sarnecka, Barbara W and Flukes, Jonathan},
  journal={Language Learning and Development},
  volume={4},
  number={1},
  pages={1--31},
  year={2008},
  publisher={Taylor \& Francis}
}

@misc{laurençon2024building,
      title={Building and better understanding vision-language models: insights and future directions.}, 
      author={Hugo Laurençon and Andrés Marafioti and Victor Sanh and Léo Tronchon},
      year={2024},
      eprint={2408.12637},
      archivePrefix={arXiv},
      primaryClass={cs.CV}
}

@inproceedings{rodriguez-etal-2025-characterizing,
    title = "Characterizing the Role of Similarity in the Property Inferences of Language Models",
    author = "Rodriguez, Juan Diego  and
      Mueller, Aaron  and
      Misra, Kanishka",
    editor = "Chiruzzo, Luis  and
      Ritter, Alan  and
      Wang, Lu",
    booktitle = "Proceedings of the 2025 Conference of the Nations of the Americas Chapter of the Association for Computational Linguistics: Human Language Technologies (Volume 1: Long Papers)",
    month = apr,
    year = "2025",
    address = "Albuquerque, New Mexico",
    publisher = "Association for Computational Linguistics",
    url = "https://aclanthology.org/2025.naacl-long.574/",
    doi = "10.18653/v1/2025.naacl-long.574",
    pages = "11515--11533",
    ISBN = "979-8-89176-189-6"
}

@article{cimpian2008preschool,
  title={Preschool children’s use of cues to generic meaning},
  author={Cimpian, Andrei and Markman, Ellen M},
  journal={Cognition},
  volume={107},
  number={1},
  pages={19--53},
  year={2008},
  publisher={Elsevier}
}

@article{comanici2025gemini,
  title={Gemini 2.5: Pushing the frontier with advanced reasoning, multimodality, long context, and next generation agentic capabilities},
  author={Comanici, Gheorghe and Bieber, Eric and Schaekermann, Mike and Pasupat, Ice and Sachdeva, Noveen and Dhillon, Inderjit and Blistein, Marcel and Ram, Ori and Zhang, Dan and Rosen, Evan and others},
  journal={arXiv preprint arXiv:2507.06261},
  year={2025}
}

@inproceedings{misra2021language,
  title={Do language models learn typicality judgments from text?},
  author={Misra, Kanishka and Ettinger, Allyson and Rayz, Julia},
  booktitle={Proceedings of the Annual Meeting of the Cognitive Science Society},
  volume={43},
  year={2021}
}

@inproceedings{misra2022property,
  title={A Property Induction Framework for Neural Language Models},
  author={Misra, Kanishka and Rayz, Julia and Ettinger, Allyson},
  booktitle={Proceedings of the Annual Meeting of the Cognitive Science Society},
  volume={44},
  year={2022}
}

@article{han2024inductive,
  title={Inductive reasoning in humans and large language models},
  author={Han, Simon Jerome and Ransom, Keith J and Perfors, Andrew and Kemp, Charles},
  journal={Cognitive Systems Research},
  volume={83},
  pages={101155},
  year={2024},
  publisher={Elsevier}
}

@inproceedings{bender-koller-2020-climbing,
    title = "Climbing towards {NLU}: {On} Meaning, Form, and Understanding in the Age of Data",
    author = "Bender, Emily M.  and
      Koller, Alexander",
    editor = "Jurafsky, Dan  and
      Chai, Joyce  and
      Schluter, Natalie  and
      Tetreault, Joel",
    booktitle = "Proceedings of the 58th Annual Meeting of the Association for Computational Linguistics",
    year = "2020",
    address = "Online",
    publisher = "Association for Computational Linguistics",
    url = "https://aclanthology.org/2020.acl-main.463/",
    doi = "10.18653/v1/2020.acl-main.463",
    pages = "5185--5198",
}

@article{mahowald2024dissociating,
  title={Dissociating language and thought in large language models},
  author={Mahowald, Kyle and Ivanova, Anna A and Blank, Idan A and Kanwisher, Nancy and Tenenbaum, Joshua B and Fedorenko, Evelina},
  year={2024},
  journal={Trends in cognitive sciences},
  publisher={Elsevier}
}

@article{osherson1990category,
  title={Category-based induction.},
  author={Osherson, Daniel N and Smith, Edward E and Wilkie, Ormond and Lopez, Alejandro and Shafir, Eldar},
  journal={Psychological review},
  volume={97},
  number={2},
  pages={185},
  year={1990},
  publisher={American Psychological Association}
}

@article{leslie2008generics,
  title={Generics: Cognition and acquisition},
  author={Leslie, Sarah-Jane},
  journal={Philosophical review},
  volume={117},
  number={1},
  pages={1--47},
  year={2008},
  publisher={Duke University Press}
}

@inproceedings{allaway-etal-2023-penguins,
    title = "{P}enguins Don{'}t Fly: Reasoning about Generics through Instantiations and Exceptions",
    author = "Allaway, Emily  and
      Hwang, Jena D.  and
      Bhagavatula, Chandra  and
      McKeown, Kathleen  and
      Downey, Doug  and
      Choi, Yejin",
    editor = "Vlachos, Andreas  and
      Augenstein, Isabelle",
    booktitle = "Proceedings of the 17th Conference of the European Chapter of the Association for Computational Linguistics",
    month = may,
    year = "2023",
    address = "Dubrovnik, Croatia",
    publisher = "Association for Computational Linguistics",
    url = "https://aclanthology.org/2023.eacl-main.192/",
    doi = "10.18653/v1/2023.eacl-main.192",
    pages = "2618--2635"
}

@inproceedings{frank-allaway-2025-visage,
    title = "{VIS}a{GE}: Understanding Visual Generics and Exceptions",
    author = "Frank, Stella  and
      Allaway, Emily",
    editor = "Christodoulopoulos, Christos  and
      Chakraborty, Tanmoy  and
      Rose, Carolyn  and
      Peng, Violet",
    booktitle = "Proceedings of the 2025 Conference on Empirical Methods in Natural Language Processing",
    month = nov,
    year = "2025",
    address = "Suzhou, China",
    publisher = "Association for Computational Linguistics",
    url = "https://aclanthology.org/2025.emnlp-main.1655/",
    doi = "10.18653/v1/2025.emnlp-main.1655",
    pages = "32537--32546",
    ISBN = "979-8-89176-332-6"
}

@inproceedings{cilleruelo-etal-2025-generics,
    title = "Generics are puzzling. Can language models find the missing piece?",
    author = "Cilleruelo, Gustavo  and
      Allaway, Emily  and
      Haddow, Barry  and
      Birch, Alexandra",
    editor = "Rambow, Owen  and
      Wanner, Leo  and
      Apidianaki, Marianna  and
      Al-Khalifa, Hend  and
      Eugenio, Barbara Di  and
      Schockaert, Steven",
    booktitle = "Proceedings of the 31st International Conference on Computational Linguistics",
    month = jan,
    year = "2025",
    address = "Abu Dhabi, UAE",
    publisher = "Association for Computational Linguistics",
    url = "https://aclanthology.org/2025.coling-main.438/",
    pages = "6571--6588"
}

@article{allaway2024exceptions,
  title={Exceptions, instantiations, and overgeneralization: Insights into how language models process generics},
  author={Allaway, Emily and Bhagavatula, Chandra and Hwang, Jena D and McKeown, Kathleen and Leslie, Sarah-Jane},
  journal={Computational Linguistics},
  volume={50},
  number={4},
  pages={1211--1275},
  year={2024},
  publisher={MIT Press 255 Main Street, 9th Floor, Cambridge, Massachusetts 02142, USA~…}
}

@inproceedings{misra-etal-2023-comps,
    title = "{COMPS}: Conceptual Minimal Pair Sentences for testing Robust Property Knowledge and its Inheritance in Pre-trained Language Models",
    author = "Misra, Kanishka  and
      Rayz, Julia  and
      Ettinger, Allyson",
    editor = "Vlachos, Andreas  and
      Augenstein, Isabelle",
    booktitle = "Proceedings of the 17th Conference of the European Chapter of the Association for Computational Linguistics",
    month = may,
    year = "2023",
    address = "Dubrovnik, Croatia",
    publisher = "Association for Computational Linguistics",
    url = "https://aclanthology.org/2023.eacl-main.213/",
    doi = "10.18653/v1/2023.eacl-main.213",
    pages = "2928--2949"
}

@inproceedings{geiger2024finding,
  title={Finding alignments between interpretable causal variables and distributed neural representations},
  author={Geiger, Atticus and Wu, Zhengxuan and Potts, Christopher and Icard, Thomas and Goodman, Noah},
  booktitle={Causal Learning and Reasoning},
  pages={160--187},
  year={2024},
  organization={PMLR}
}

@article{lakretz2021mechanisms,
  title={Mechanisms for handling nested dependencies in neural-network language models and humans},
  author={Lakretz, Yair and Hupkes, Dieuwke and Vergallito, Alessandra and Marelli, Marco and Baroni, Marco and Dehaene, Stanislas},
  journal={Cognition},
  volume={213},
  pages={104699},
  year={2021},
  publisher={Elsevier}
}

@article{mccloskey1991networks,
  title={Networks and theories: The place of connectionism in cognitive science},
  author={McCloskey, Michael},
  journal={Psychological science},
  volume={2},
  number={6},
  pages={387--395},
  year={1991},
  publisher={SAGE Publications Sage CA: Los Angeles, CA}
}

@article{misra2024generating,
  title={Generating novel experimental hypotheses from language models: A case study on cross-dative generalization},
  author={Misra, Kanishka and Kim, Najoung},
  journal={arXiv preprint arXiv:2408.05086},
  year={2024}
}

@article{cao2024contra,
  title={Explanatory models in neuroscience, Part 2: Functional intelligibility and the contravariance principle},
  author={Cao, Rosa and Yamins, Daniel},
  journal={Cognitive Systems Research},
  volume={85},
  pages={101200},
  year={2024},
  publisher={Elsevier}
}

@article{futrell2025linguistics,
  title={How linguistics learned to stop worrying and love the language models},
  author={Futrell, Richard and Mahowald, Kyle},
  journal={arXiv preprint arXiv:2501.17047},
  year={2025}
}

@article{cao2024explanatory,
  title={Explanatory models in neuroscience, Part 1: Taking mechanistic abstraction seriously},
  author={Cao, Rosa and Yamins, Daniel},
  journal={Cognitive Systems Research},
  volume={87},
  pages={101244},
  year={2024},
  publisher={Elsevier}
}

@article{lampinen2025representation,
  title={Representation biases: will we achieve complete understanding by analyzing representations?},
  author={Lampinen, Andrew Kyle and Chan, Stephanie CY and Li, Yuxuan and Hermann, Katherine},
  journal={arXiv preprint arXiv:2507.22216},
  year={2025}
}

@inproceedings{
yiu2025kiva,
title={Ki{VA}: Kid-inspired Visual Analogies for Testing Large Multimodal Models},
author={Eunice Yiu and Maan Qraitem and Anisa Noor Majhi and Charlie Wong and Yutong Bai and Shiry Ginosar and Alison Gopnik and Kate Saenko},
booktitle={The Thirteenth International Conference on Learning Representations},
year={2025},
url={https://openreview.net/forum?id=vNATZfmY6R}
}

@inproceedings{
tan2024devbench,
title={DevBench: A multimodal developmental benchmark for language learning},
author={Alvin Wei Ming Tan and Chunhua Yu and Bria Lorelle Long and Wanjing Anya Ma and Tonya Murray and Rebecca D. Silverman and Jason D Yeatman and Michael Frank},
booktitle={The Thirty-eight Conference on Neural Information Processing Systems Datasets and Benchmarks Track},
year={2024},
url={https://openreview.net/forum?id=zogaeVpbaE}
}

@article{firestone2020performance,
  title={Performance vs. competence in human--machine comparisons},
  author={Firestone, Chaz},
  journal={Proceedings of the National Academy of Sciences},
  volume={117},
  number={43},
  pages={26562--26571},
  year={2020},
  publisher={National Academy of Sciences}
}

@inproceedings{
qin2025visionandlanguage,
title={Vision-and-Language Training Helps Deploy Taxonomic Knowledge but Does Not Fundamentally Alter It},
author={Yulu Qin and Dheeraj Varghese and Adam Dahlgren Lindstr{\"o}m and Lucia Donatelli and Kanishka Misra and Najoung Kim},
booktitle={The Thirty-ninth Annual Conference on Neural Information Processing Systems},
year={2025},
url={https://openreview.net/forum?id=KXmDTGKwhy}
}

@inproceedings{collacciani-etal-2024-quantifying,
    title = "Quantifying Generalizations: Exploring the Divide Between Human and {LLM}s' Sensitivity to Quantification",
    author = "Collacciani, Claudia  and
      Rambelli, Giulia  and
      Bolognesi, Marianna",
    editor = "Ku, Lun-Wei  and
      Martins, Andre  and
      Srikumar, Vivek",
    booktitle = "Proceedings of the 62nd Annual Meeting of the Association for Computational Linguistics (Volume 1: Long Papers)",
    month = aug,
    year = "2024",
    address = "Bangkok, Thailand",
    publisher = "Association for Computational Linguistics",
    url = "https://aclanthology.org/2024.acl-long.636/",
    doi = "10.18653/v1/2024.acl-long.636",
    pages = "11811--11822"
}

@inproceedings{ralethe-buys-2022-generic,
    title = "Generic Overgeneralization in Pre-trained Language Models",
    author = "Ralethe, Sello  and
      Buys, Jan",
    editor = "Calzolari, Nicoletta  and
      Huang, Chu-Ren  and
      Kim, Hansaem  and
      Pustejovsky, James  and
      Wanner, Leo  and
      Choi, Key-Sun  and
      Ryu, Pum-Mo  and
      Chen, Hsin-Hsi  and
      Donatelli, Lucia  and
      Ji, Heng  and
      Kurohashi, Sadao  and
      Paggio, Patrizia  and
      Xue, Nianwen  and
      Kim, Seokhwan  and
      Hahm, Younggyun  and
      He, Zhong  and
      Lee, Tony Kyungil  and
      Santus, Enrico  and
      Bond, Francis  and
      Na, Seung-Hoon",
    booktitle = "Proceedings of the 29th International Conference on Computational Linguistics",
    month = oct,
    year = "2022",
    address = "Gyeongju, Republic of Korea",
    publisher = "International Committee on Computational Linguistics",
    url = "https://aclanthology.org/2022.coling-1.282/",
    pages = "3187--3196"
}

@book{murphy2004big,
  title={The big book of concepts},
  author={Murphy, Gregory},
  year={2004},
  publisher={MIT press}
}

@article{bhatia2023inductive,
  title={Inductive reasoning in minds and machines.},
  author={Bhatia, Sudeep},
  journal={Psychological Review},
  year={2023},
  publisher={American Psychological Association}
}

@article{wong2025vaquum,
  title={VAQUUM: Are Vague Quantifiers Grounded in Visual Data?},
  author={Wong, Hugh Mee and Nouwen, Rick and Gatt, Albert},
  journal={arXiv preprint arXiv:2502.11874},
  year={2025}
}

@article{marafioti2025smolvlm,
  title={SmolVLM: Redefining small and efficient multimodal models}, 
  author={Andrés Marafioti and Orr Zohar and Miquel Farré and Merve Noyan and Elie Bakouch and Pedro Cuenca and Cyril Zakka and Loubna Ben Allal and Anton Lozhkov and Nouamane Tazi and Vaibhav Srivastav and Joshua Lochner and Hugo Larcher and Mathieu Morlon and Lewis Tunstall and Leandro von Werra and Thomas Wolf},
  journal={arXiv preprint arXiv:2504.05299},
  year={2025}
}

@inproceedings{liu2023llava,
    author      = {Liu, Haotian and Li, Chunyuan and Wu, Qingyang and Lee, Yong Jae},
    title       = {Visual Instruction Tuning},
    booktitle   = {NeurIPS},
    year        = {2023}
  }

@misc{liu2023improvedllava,
          author={Liu, Haotian and Li, Chunyuan and Li, Yuheng and Lee, Yong Jae},
          title={Improved Baselines with Visual Instruction Tuning}, 
          publisher={arXiv:2310.03744},
          year={2023},
  }

@inproceedings{deitke2025molmo,
  title={Molmo and pixmo: Open weights and open data for state-of-the-art vision-language models},
  author={Deitke, Matt and Clark, Christopher and Lee, Sangho and Tripathi, Rohun and Yang, Yue and Park, Jae Sung and Salehi, Mohammadreza and Muennighoff, Niklas and Lo, Kyle and Soldaini, Luca and others},
  booktitle={Proceedings of the Computer Vision and Pattern Recognition Conference},
  pages={91--104},
  year={2025}
}

@misc{qwen2.5-VL,
    title = {Qwen2.5-VL},
    url = {https://qwenlm.github.io/blog/qwen2.5-vl/},
    author = {{Qwen Team}},
    month = {January},
    year = {2025}
}

@misc{qwen3technicalreport,
      title={Qwen3 Technical Report}, 
      author={{Qwen Team}},
      year={2025},
      eprint={2505.09388},
      archivePrefix={arXiv},
      primaryClass={cs.CL},
      url={https://arxiv.org/abs/2505.09388}, 
}

\appendix

\section{Selected Models}
\label{sec:metadata}

\Cref{tab:models_with_hf_identifiers} shows the metadata of the models used in this paper. Models were run on a combination of NVIDIA A40 and GH100 GPUs.

\section{Data Release}

We used images from the THINGS database for our vision+language stimuli \citep{hebart2019things, THINGSdata}. They release their data with the Attribution CC BY license.\footnote{See \url{https://osf.io/jum2f/files/52wrx}} When we release our stimuli we will point to their file names and not release their data as part of our data. All our data will be released under the MIT License.

\begin{table*}[!t]
  \centering
  \footnotesize
  \renewcommand{\arraystretch}{1.2}
  \setlength{\tabcolsep}{5pt}

  \begin{tabular}{l c c p{6.4cm}}
    \toprule
    \textbf{Model} & \textbf{Params} & \textbf{Instruction Tuned} & \textbf{HF Identifier} \\
    \midrule
    SmolVLM                 & 2.2B & \checkmark & \texttt{SmolVLM-Instruct} \\
    LLaVA-1.5               & 7B   & \checkmark & \texttt{llava-1.5-7b-hf} \\
    LLaVA-1.6 (Mistral)     & 7B   & \checkmark & \texttt{llava-v1.6-mistral-7b-hf} \\
    LLaVA-OneVision (Qwen)  & 7B   & \checkmark & \texttt{llava-hf/llava-onevision-qwen2-7b-ov-hf} \\
    Molmo                   & 7B   & \checkmark & \texttt{Molmo-7B-D-0924} \\
    Qwen2.5-VL              & 7B   & \checkmark & \texttt{Qwen2.5-VL-7B-Instruct} \\
    Qwen3-VL                & 2B   & \checkmark & \texttt{Qwen3-VL-2B-Instruct} \\
    Qwen3-VL                & 4B   & \checkmark & \texttt{Qwen3-VL-4B-Instruct} \\
    Qwen3-VL                & 8B   & \checkmark & \texttt{Qwen3-VL-8B-Instruct} \\
    \bottomrule
  \end{tabular}

  \caption{Overview of models used in our experiments.}
  \label{tab:models_with_hf_identifiers}
\end{table*}

\section{Vision + Language Stimuli Generation for \textit{all} and \textit{some}}
\label{sec:all-some-generation}

\subsection{Prompts}
We prompt Gemini 2.5 Flash Image~\citep{comanici2025gemini} to first generate an image, and then prompt the model to modify the generated image. All in all, this cost around \$17 in google cloud credits.
In the first condition, we first ask the model to generate an image with all the objects in the image are in the same color, and then modify the image by adding some objects in a different color:

\begin{lstlisting}[basicstyle=\ttfamily\footnotesize, breaklines=true, columns=fullflexible, keepspaces=true, showstringspaces=false]

1: 

USER:
Create a picture of {number_majority} {color} {object}. Make sure the background is as clean as possible.

MODEL: [IMAGE_ALL]
\end{lstlisting}

\begin{lstlisting}[basicstyle=\ttfamily\footnotesize, breaklines=true, columns=fullflexible, keepspaces=true, showstringspaces=false]

2:

USER:
[IMAGE_ALL] Modify this picture by adding {number_minority} {color} {object}. Change nothing else.

MODEL:
[IMAGE_SOME]
\end{lstlisting}

In the second condition, we first prompt the model to generate an image with all the objects in the image in a container, and then prompt it to modify the image by adding some objects (minority) out of the container:

\begin{lstlisting}[basicstyle=\ttfamily\footnotesize, breaklines=true, columns=fullflexible, keepspaces=true, showstringspaces=false]

1:

USER:
Generate a picture of {number_majority} {object} in a {container}. Make sure the background is as clean as possible.

MODEL: [IMAGE_ALL]
\end{lstlisting}

\begin{lstlisting}[basicstyle=\ttfamily\footnotesize, breaklines=true, columns=fullflexible, keepspaces=true, showstringspaces=false]
2:

USER:
[IMAGE_ALL] Modify this picture by adding {number_minority} {object} outside the {container}. Change nothing else.

MODEL:
[IMAGE_SOME]
\end{lstlisting}

In the third condition, we first prompt the model to generate an image with most of the objects (majority) in a container, and some of them (minority) out of it. Then we prompt the model to modify the image by removing the objects that are out of the container.
\begin{lstlisting}[basicstyle=\ttfamily\footnotesize, breaklines=true, columns=fullflexible, keepspaces=true, showstringspaces=false]

1:

USER:
Generate a picture of {number_majority} {object} in a {container} and {number_minority} out of it. Make sure the background is as clean as possible.

MODEL: [IMAGE_SOME]
\end{lstlisting}

\begin{lstlisting}[basicstyle=\ttfamily\footnotesize, breaklines=true, columns=fullflexible, keepspaces=true, showstringspaces=false]
2:

USER:
[IMAGE_SOME] Modify this picture by removing the {number_minority} {object} outside the {container}. Change nothing else.

MODEL:
[IMAGE_ALL]
\end{lstlisting}

\subsection{Annotation}
\begin{figure*}
    \centering
    \includegraphics[width=\linewidth]{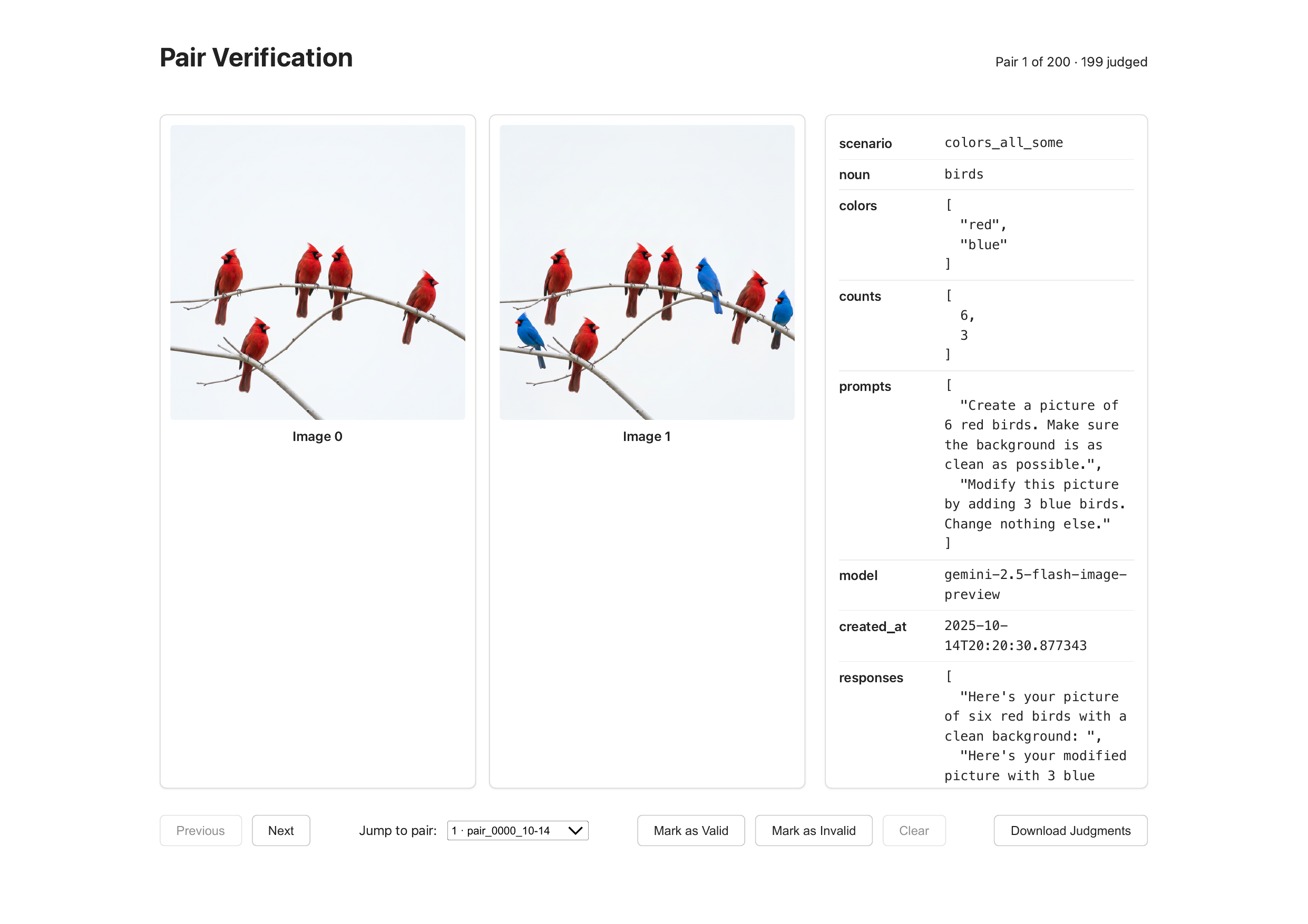}
    \caption{Frontend for annotating image pairs generated by Gemini-2.5}
    \label{fig:frontend-screenshot}
\end{figure*}

\begin{figure*}[!t]
  \centering

  % Row 1
  \begin{subfigure}[t]{0.48\linewidth}
    \centering
    \includegraphics[width=\linewidth]{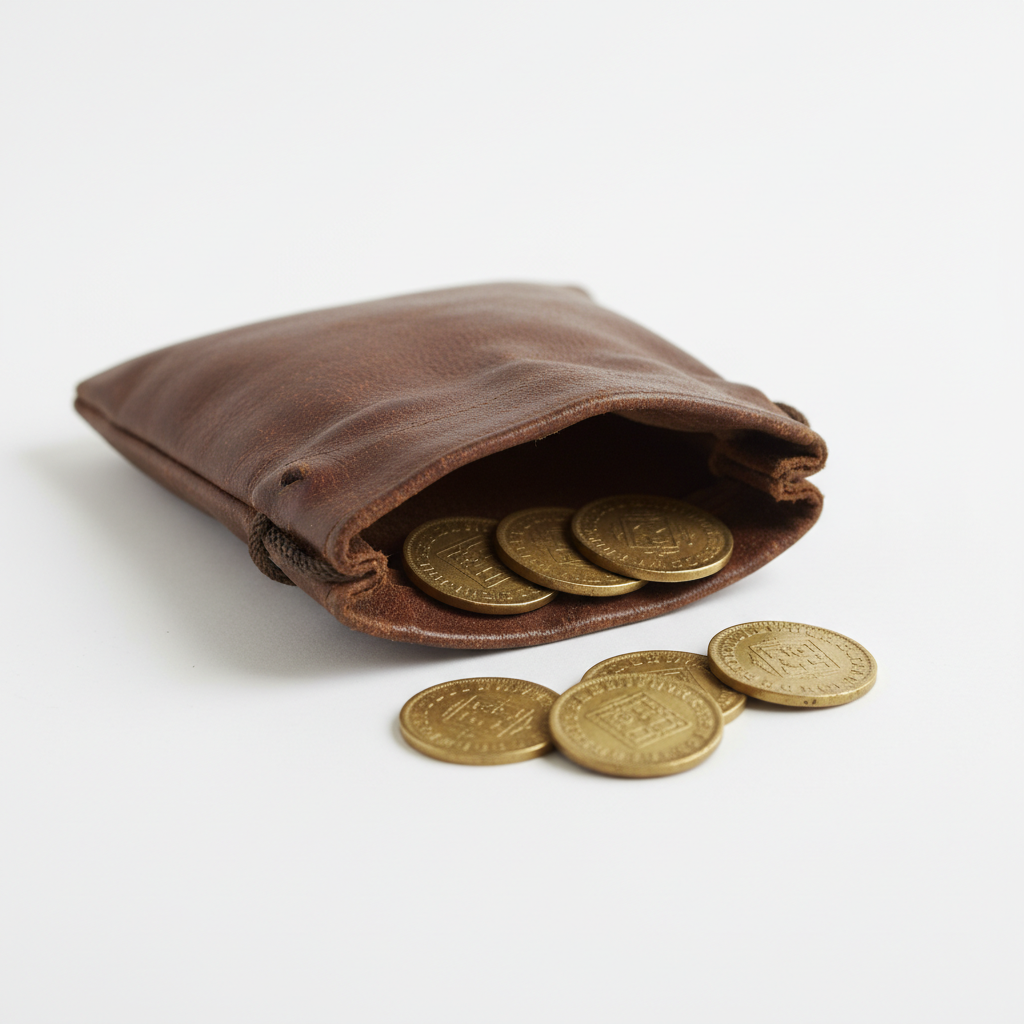}
    \caption*{(1a)}
    \label{fig:1a}
  \end{subfigure}
  \hfill
  \begin{subfigure}[t]{0.48\linewidth}
    \centering
    \includegraphics[width=\linewidth]{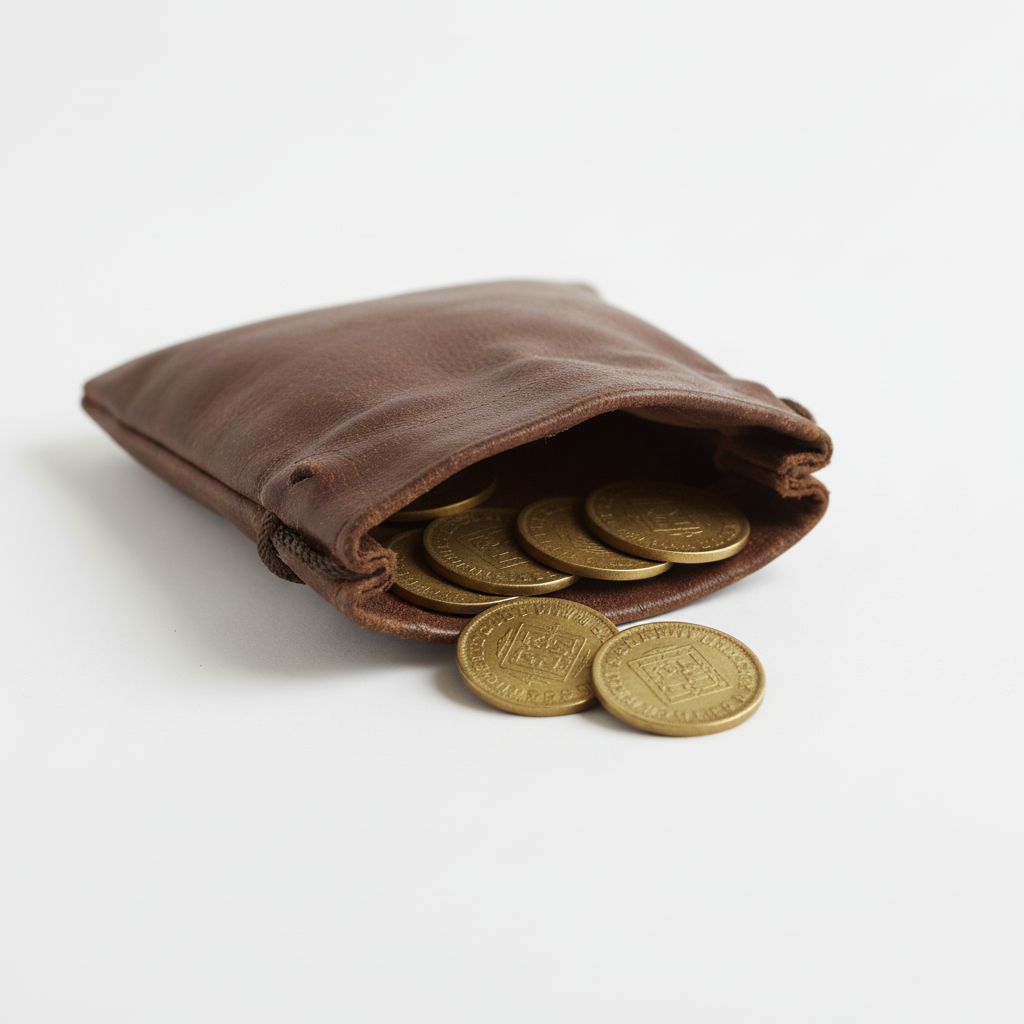}
    \caption*{(1b)}
    \label{fig:1b}
  \end{subfigure}

  \vspace{0.25em}
  %\caption*{Caption}

  \vspace{0.8em}

  % Row 2
  \begin{subfigure}[t]{0.48\linewidth}
    \centering
    \includegraphics[width=\linewidth]{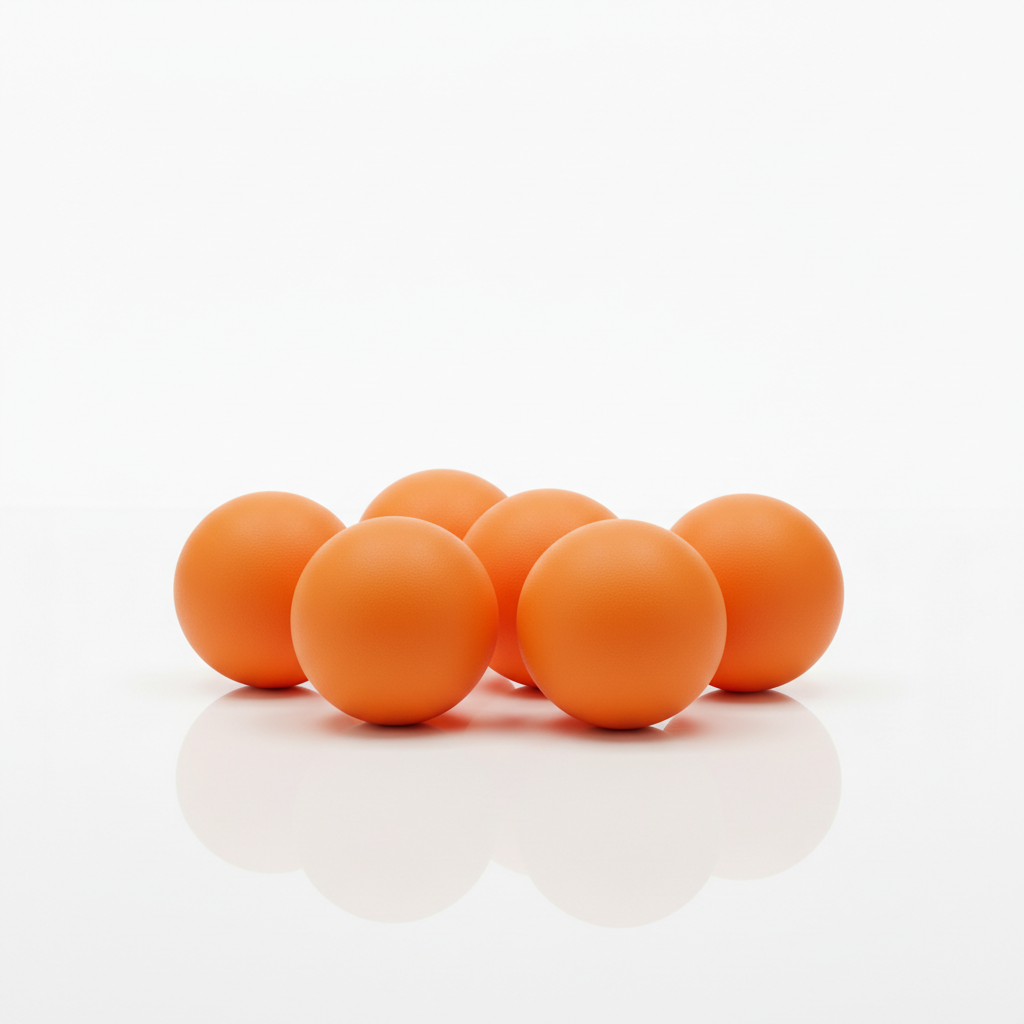}
    \caption*{(2a)}
    \label{fig:2a}
  \end{subfigure}
  \hfill
  \begin{subfigure}[t]{0.48\linewidth}
    \centering
    \includegraphics[width=\linewidth]{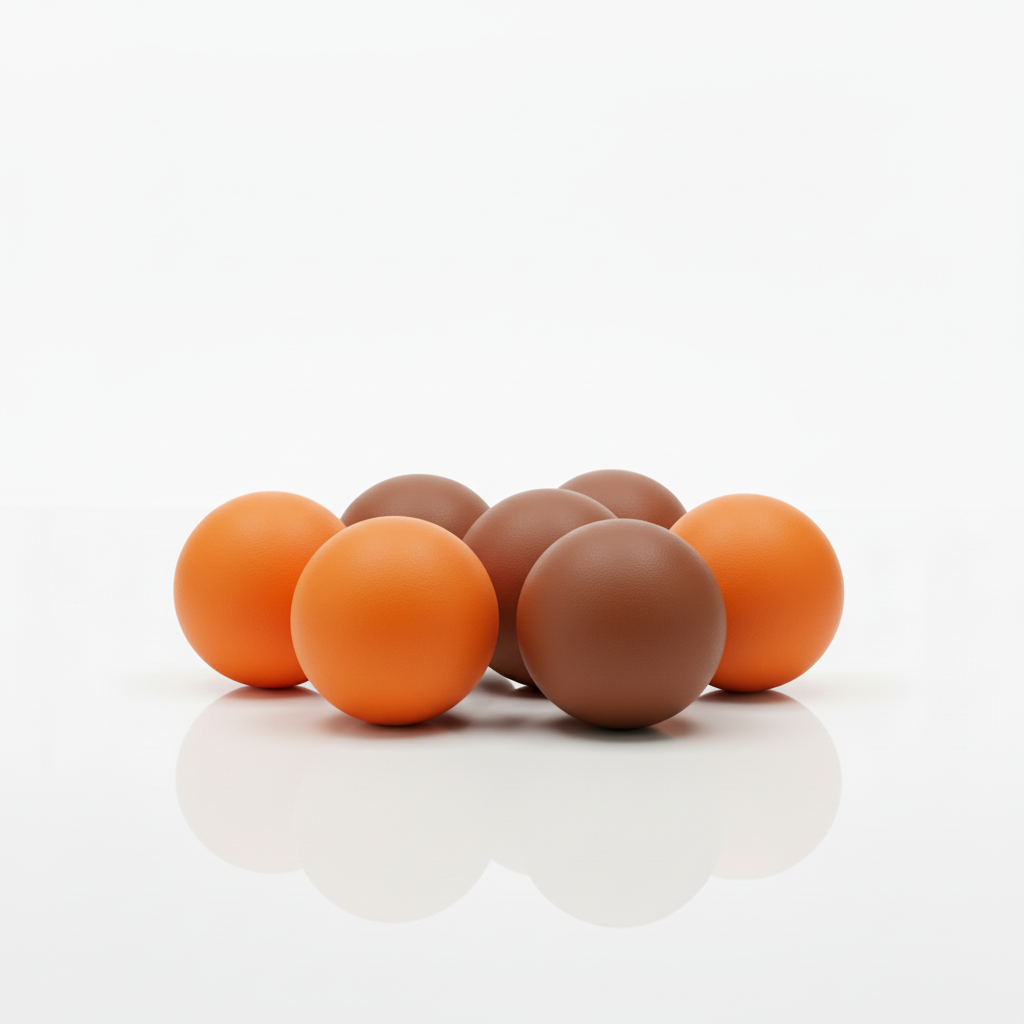}
    \caption*{(2b)}
    \label{fig:2b}
  \end{subfigure}

  \vspace{0.25em}
 %\caption*{Caption 2.}

  \vspace{0.8em}

  \caption{Examples of invalid image pairs. For the first pair, the model was prompted to generate an image with six coins in the bag and three coins outside the bag in (1a), and then to remove the three coins outside the bag in (1b); however, the model did not follow the prompt, and 2 coins still appear outside the bag in (1b). For the second pair, the model was prompted to generate an image with 6 orange balls in (2a), and then to add three brown balls in (2b); however, the model changed the color of balls when prompted to modify the image.}
  \label{fig:invalid_pairs}
\end{figure*}

We manually annotate the 200 generated image pairs using the frontend shown in Figure \ref{fig:frontend-screenshot}. We excluded 60 pairs from our dataset. Examples of excluded pairs are shown in Figure \ref{fig:invalid_pairs}. 

\section{Surface form variation in the inductive generalization stimuli}
\label{appsec:surface-form}

In order to robustly identify how tested VLMs differentiate between \textit{generics}, \textit{all} and \textit{some}, we use 4 different surface form variations for each stimuli type-category combination. Table 4 shows the different surface form variations used for the stimuli in Experiment 3 along with examples of each. 

\begin{table*}[!t]
  \centering
  \footnotesize
  \renewcommand{\arraystretch}{1.2}
  \setlength{\tabcolsep}{5pt}

  \begin{tabular}{l p{0.75\linewidth}}
    \toprule
    \textbf{Template Name} &\textbf{Prompt}\\
    \toprule
    Given that & Given that all bears have the T9 Hormone, does this bear have the T9 Hormone?\\
    Suppose & Suppose all bears have the T9 Hormone. Does this bear have the T9 Hormone?\\
    If & If all bears have the T9 Hormone, does this bear have the T9 Hormone?\\
    No Template & All bears have the T9 Hormone. Does this bear have the T9 Hormone?\\
    \bottomrule
  \end{tabular}

  \caption{Surface Form Variations for Experiment 3}
  \label{tab:surface_form_variation}
\end{table*}

\begin{table*}[!t]
  \centering
  \footnotesize
  \renewcommand{\arraystretch}{1.2}

  \begin{tabular}{l l p{0.55\linewidth}}
    \toprule
    \textbf{Type} & \textbf{Animacy} & \textbf{Property} \\
    \midrule
    \multirow{2}{*}{Real-sounding} & Animate & has the T9 Hormone / has Interleukin 54 / has Homoglobin 76 / has Benzokinase / has Immunoglobulin P \\
                        & Inanimate & is made of bergentium, is made of misrium, is heteromagnetic, is betamagnetic, contains ostinium \\
    Nonce & Animate/Inanimate & is/are daxable, has/have feps, can dax, can gek, is/are tomable \\
    \bottomrule
  \end{tabular}

  \caption{Properties for Experiment 3. The nonce properties are used universally for Animate and Inanimate concepts.}
  \label{tab:your_label}
\end{table*}

\section{Detailed PCA results}
\label{appsec:pca-full}

\begin{figure*}
    \centering
    \includegraphics[width=\linewidth]{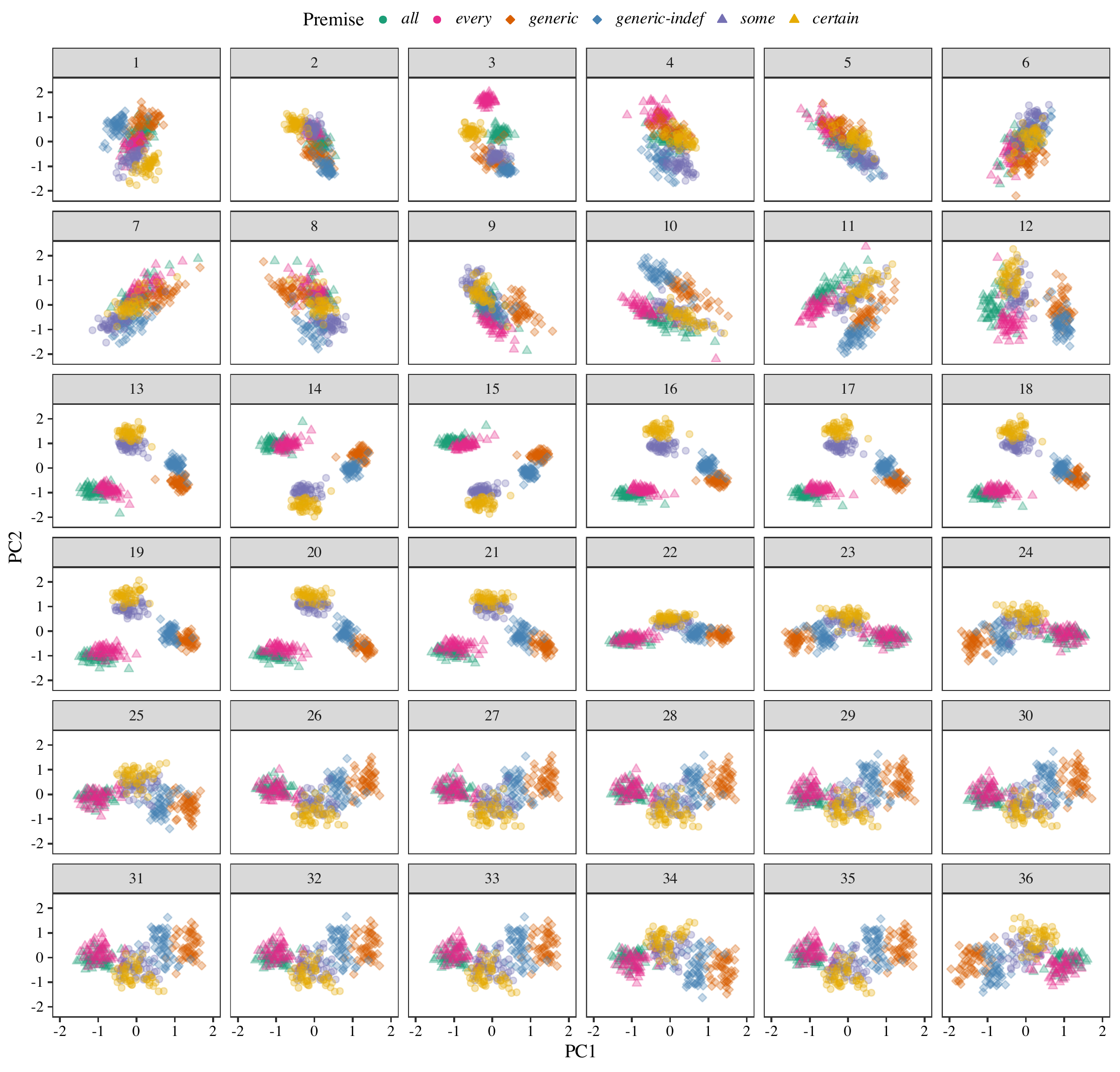}
    \caption{First and second Principal Components of the last hidden state representations in all layers of Qwen3-VL-8B for stimuli that attribute properties to categories and vary in their scope, as modulated by quantifiers (all/every vs. some/certain) or generics (bare plural/indefinite).}
    \label{fig:pca-full}
\end{figure*}

\Cref{fig:pca-full} shows low dimensional representations of the stimuli in our post-hoc analysis in \Cref{sec:main-exp-results}. We see strong evidence of inductive-behavior-based separation of points starting to emerge at layer 13. To further quantify the degree of closeness between classes of stimuli, we compute the average pairwise euclidean distances of each point to every other point, and visualize results by taking the average euclidean distance between the set of points within each type of premise. We see that the average distance between points belonging to propositions with similar inductive constraints (\textit{generic--generic-indef}, \textit{all--every}, \textit{some-certain}) is lowest, relative to all others, and this pattern emerges around layer 13, and becomes stable at around layer 22.

\begin{figure*}
    \centering
    \includegraphics[width=\linewidth]{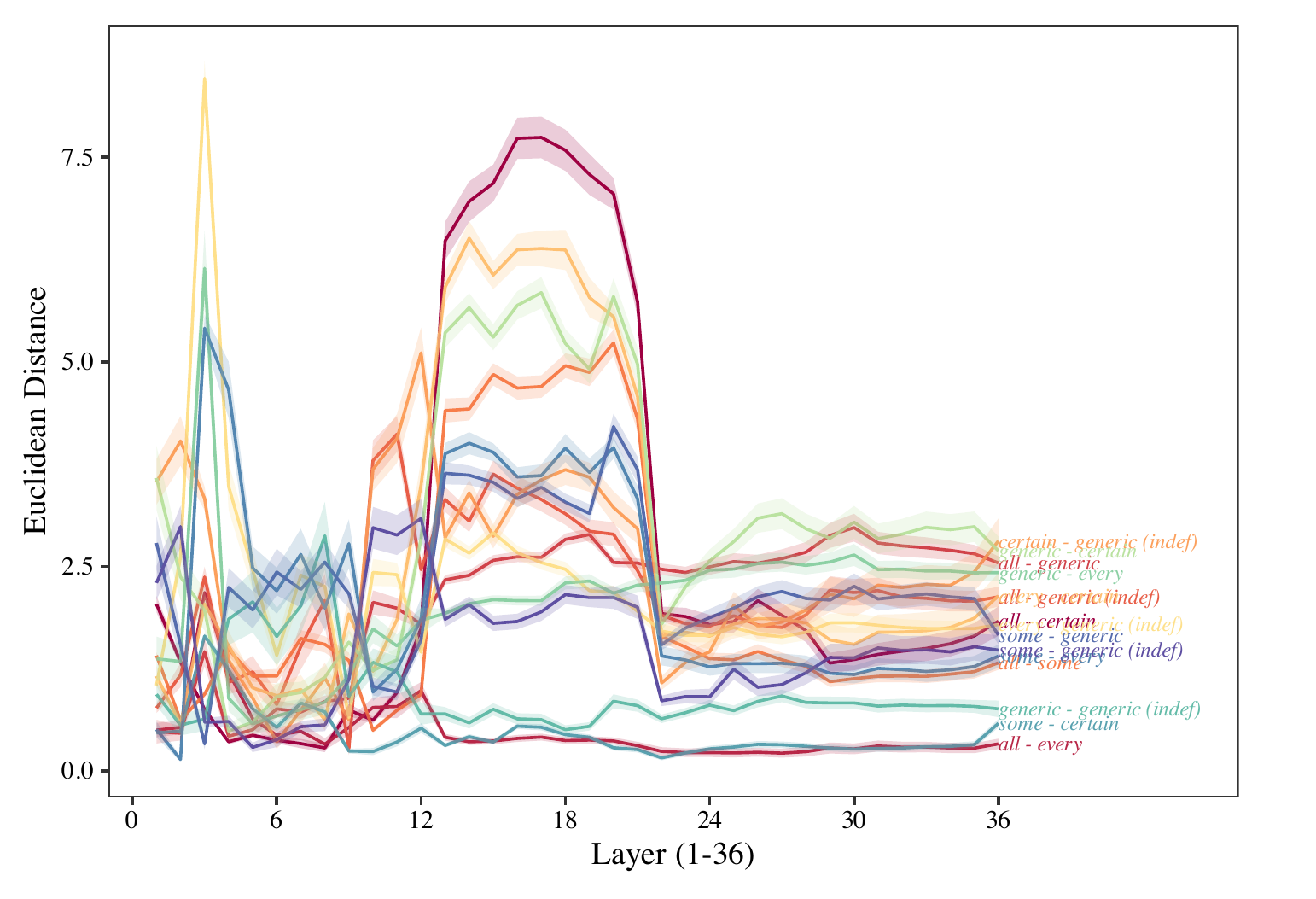}
    \caption{Average euclidean distance between collection of points reduced to two dimensions (using PCA). The average distance between points belonging to propositions with similar inductive constraints (\textit{generic--generic-indef}, \textit{all--every}, \textit{some-certain}) is lowest, relative to all others, and those emerges around layer 13.}
    \label{fig:placeholder}
\end{figure*}

\end{document}